\documentclass[11pt]{article}

\usepackage[final]{acl}

\usepackage{times}
\usepackage{latexsym}
\usepackage{multirow}

\usepackage[T1]{fontenc}

\usepackage[utf8]{inputenc}

\usepackage{microtype}

\usepackage{inconsolata}

\usepackage{graphicx}

\usepackage{amsmath,amsfonts}
\usepackage{algorithmic}
\usepackage{algorithm}

\usepackage{array}
\usepackage{textcomp}
\usepackage{url}
\usepackage{verbatim}
\usepackage{subcaption}
\usepackage{booktabs}

\usepackage{amsthm} 
\usepackage{balance}
\usepackage{xcolor} 

\newif\ifarxivversion
\arxivversionfalse

\newtheorem{theorem}{Theorem}

\newtheorem{proposition}{Proposition}

\title{TPA: Next Token Probability Attribution for \\ Detecting Hallucinations in RAG}

\author{Pengqian Lu, Jie Lu\thanks{Corresponding author.}, Anjin Liu, \and Guangquan Zhang \\
  Australian Artificial Intelligence Institute (AAII) \\
  University of Technology Sydney \\
  Ultimo, NSW 2007, Australia \\
  \texttt{\{Pengqian.Lu@student., Jie.Lu@, Anjin.Liu@, Guangquan.Zhang@\}uts.edu.au}}

\begin{document}
\maketitle
\begin{abstract}
Detecting hallucinations in Retrieval-Augmented Generation (RAG) remains a critical reliability challenge, as ungrounded responses can have severe consequences in high-stakes applications such as clinical decision support, legal research assistants, and autonomous agents that act on retrieved evidence. Prior approaches  attribute hallucinations to a binary conflict between internal knowledge stored in FFNs  and the retrieved context. However, this perspective is incomplete, failing to account for the impact of other components of the LLM, such as the user query, previously generated tokens, the self token, and the Final LayerNorm adjustment. To comprehensively capture the impact of these components on hallucination detection, we propose TPA which mathematically attributes each token's probability to seven distinct sources: Query, RAG Context, Past Token, Self Token, FFN, Final LayerNorm, and Initial Embedding. This attribution  quantifies how each source contributes to the generation of the next token. Specifically, we aggregate these attribution scores by Part-of-Speech (POS) tags to quantify the contribution of each model component to the generation of specific linguistic categories within a response. By leveraging these patterns, such as detecting anomalies where Nouns rely heavily on LayerNorm, TPA effectively identifies hallucinated responses. Extensive experiments on five LLMs (Llama2-7B/13B, Llama3-8B, Mistral-7B, and Qwen3-8B) demonstrate that TPA achieves state-of-the-art performance across diverse architectures.
\end{abstract}

\begin{figure}[!t]
    \centering
    \begin{subfigure}{\linewidth}
        \centering
        \includegraphics[width=\linewidth]{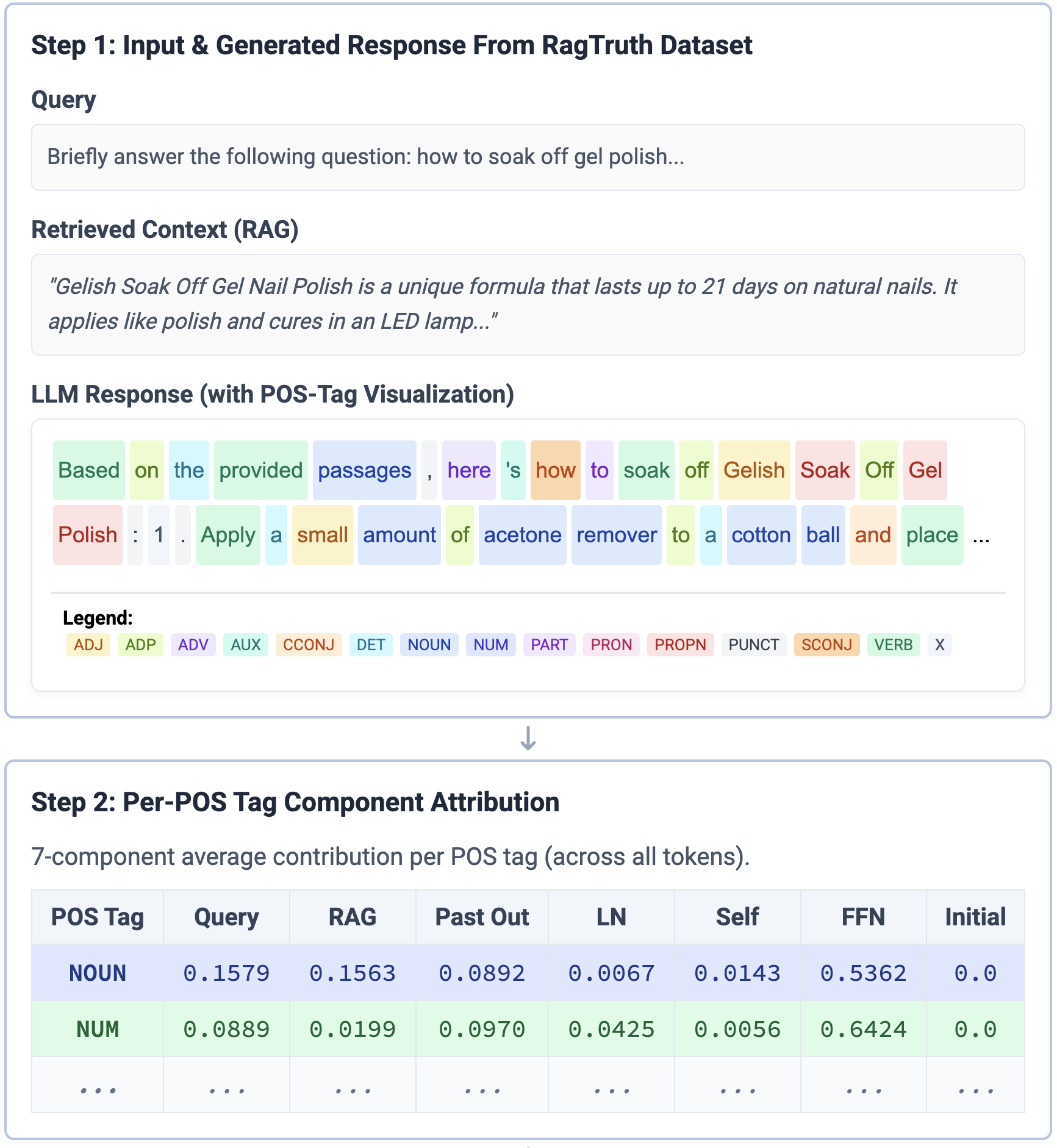}
        \caption{TPA attributes each next-token probability to seven sources, then aggregates source attributions by POS tags to form the features for detecting hallucination.}
        \label{fig:intro-fig-a}
    \end{subfigure}
    
    \vspace{0.2cm}
    
    \begin{subfigure}{\linewidth}
        \centering
        \includegraphics[width=1\linewidth]{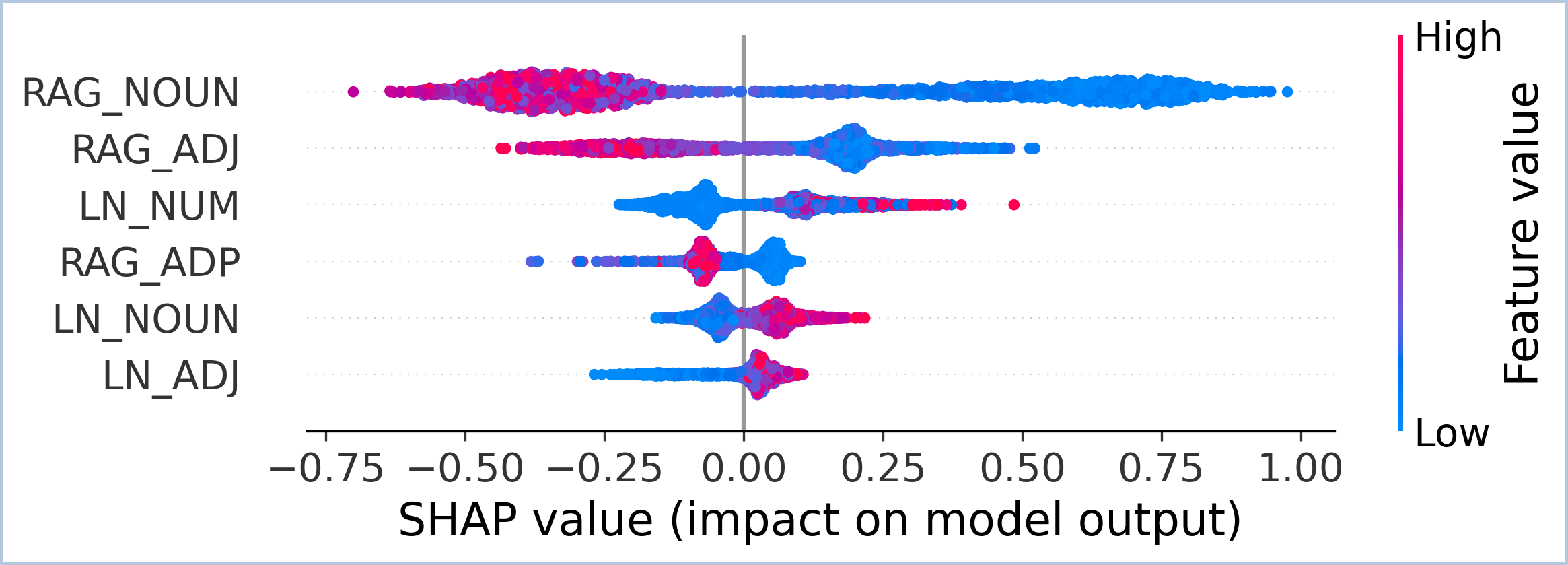}
        \caption{ Feature-importance analysis (SHAP) shows that the detector leverages {source contributions conditioned on POS tags}. 
        For example, responses are more likely to be hallucinated when {RAG} contributes little to \texttt{NOUN} tokens or when {LN} contributes too much to \texttt{NUM} tokens.}

        \label{fig:intro-fig-b}
    \end{subfigure}
    
    \caption{Applying the TPA framework to a Llama2-7b response from RAGTruth dataset \cite{niu2024ragtruth}.
    }
    \label{fig:into-fig}
\end{figure}
\section{Introduction}
Large Language Models (LLMs), despite their impressive capabilities, are prone to hallucinations \cite{huang2025survey}. Consequently, Retrieval-Augmented Generation (RAG) \cite{rag} is widely used to alleviate hallucinations by grounding models in external knowledge. However, RAG systems are not perfect. They can still hallucinate by ignoring or misinterpreting the retrieved information \cite{sun2024redeep}. Detecting such failures is therefore a critical challenge. Following \citet{sun2024redeep}, we define a \emph{hallucination} as a response containing content inconsistent with the retrieved RAG context (assuming the context is relevant and correct), excluding retrieval errors, outdated knowledge, and ambiguous evidence. The cost rises with stakes: hallucinated medication dosages in clinical decision support can harm patients, fabricated case citations in legal research assistants \cite{lu2026querycounselstructuredreasoning} have led to sanctioned court filings, and ungrounded intermediate responses in autonomous agents \cite{lu2026choosing} propagate into downstream actions.


The prevailing paradigm for hallucination detection typically relies on hand-crafted proxy signals. 
For example, common approaches detect hallucination through consistency checks~\cite{manakul2023selfcheckgpt} 
or scalar uncertainty metrics such as semantic entropy~\cite{han2024semantic}. 
However, these methods only measure the {symptoms} of hallucination, such as output variance or surface confidence, 
rather than the underlying architectural causes. Consequently, they often fail when a model is confidently incorrect~\cite{simhi2025trust}.

To address the root cause of hallucination, recent research has shifted focus to the model's internal representations. Pioneering works such as ReDeEP \cite{sun2024redeep} {explicitly assume} the RAG context is correct. They reveal that hallucinations in RAG typically stem from a {disproportionate} dominance of internal parametric knowledge (stored in FFNs) over the retrieved external context.

This insight inspires a fundamental question: \textbf{Is the \textbf{binary} conflict between FFNs and RAG the only cause of hallucination?} Critical components like LayerNorm and User Query are often overlooked. \textbf{Do contributions from these sources also drive hallucinations?} In this paper, we extend the analysis to cover all additive components along the transformer residual stream. 
This approach enables detection based on the model's full internal mechanics instead of relying on partial proxy signals.

To achieve this, we also assume the RAG context contains relevant information and introduce TPA (Next \textbf{T}oken \textbf{P}robability \textbf{A}ttribution for Detecting Hallucinations in RAG). This framework mathematically attributes the final probability of each token to seven distinct sources: Query, RAG, Past, Self Token, FFN, Final LayerNorm, and Initial Embedding. The attribution scores of these seven parts sum to the token's final probability, ensuring we capture the complete generation process.

To compute these attributions, we proposed a probe function similar with \citet{logitlens} that uses the model's unembedding matrix to read out the next-token probability directly from an intermediate residual-stream state.
Concretely, for each component on the residual stream, we define its {contribution} as the change in the probed next-token probability \emph{before} versus \emph{after} applying that component. 
In this way, we can compute the contribution from Initial Embedding, attention block, FFN, and Final LayerNorm.
For the attention block, we further distribute its contribution to Query, RAG, Past and Self Token according to their attention weights.

However, these attention scores are insufficient for detection. A high reliance on internal parametric knowledge (FFNs) does not necessarily imply a hallucination. This pattern is expected for function words like "the" or "of". Yet, it becomes highly suspicious when found in named entities. Therefore, treating all tokens equally fails to capture these critical distinctions.

To capture this distinction, we aggregate the attribution scores using Part-of-Speech (POS) tags. We employ POS tags to capture comprehensive syntactic patterns. Unlike Named Entity Recognition (NER), which is limited to specific entity types, POS tagging covers all tokens (including critical categories like Numerals and Adpositions) and maintains high computational efficiency.

Figure~\ref{fig:into-fig} illustrates how TPA turns a single response into detection features: we first compute token-level source attributions, then aggregate them by POS tags. The second step is critical since hallucination signals vary across distinct parts of speech.
For example, low {RAG} contribution on nouns or high {LN} contribution on numerals is often indicative of hallucination. These patterns are harder to capture if we only use raw token-level attribution scores without POS information.

Our main contributions are:
\begin{enumerate} 
\item We propose TPA, a novel framework that mathematically attributes each token's probability to seven distinct attribution sources. This provides a comprehensive mechanistic view of the token generation process.
\item We introduce a syntax-aware aggregation mechanism. By quantifying how attribution sources drive distinct parts of speech, this approach enables the detector to pinpoint anomalies in specific entities while ignoring benign grammatical patterns.
\item Extensive experiments demonstrate that TPA achieves state-of-the-art performance. Our framework also offers transparent interpretability, automatically uncovering novel mechanistic signatures, such as anomalous LayerNorm contributions, that extend beyond the traditional FFN-RAG binary conflict. \end{enumerate}

\section{Related Work}\label{sec:related}

\paragraph{Uncertainty and Proxy Metrics.} 
Approaches in this category estimate hallucination via output inconsistency or proxy signals.
Some methods quantify uncertainty using model ensembles \cite{malinin2021structured} or by measuring self-consistency across multiple sampled generations from a single model \cite{manakul2023selfcheckgpt}. 
Others utilize lightweight proxy scores computable from a single generation pass, such as energy-based OOD proxy scores \cite{liu2020energy}, distributional prototype learning that models each in-distribution class with class-conditioned continuous distributions for OOD detection~\cite{peng2025distributional}, embedding-based distance scores for conditional LMs \cite{renout}, and token-level uncertainty heuristics for hallucination detection \cite{lee2024ced, zhang2023enhancing}. While efficient, these scores provide indirect signals (e.g., confidence or distribution shift) and therefore may be imperfect indicators of factual correctness.

\paragraph{Distribution Shift and Predictive-Distribution Modeling.}
Hallucination can also be viewed as a failure under distribution shift. Most relevant, knowledge distillation with auxiliary variables unifies logit- and feature-level predictive-distribution matching~\cite{peng2024knowledge}. Related distribution-shift lines include bias-aware prediction under long-tailed regimes~\cite{lu2025newborn}, online adaptation under concept drift~\cite{yu2024online, yu2026drift, yu2026generalized}, and structural representation learning such as deep subspace clustering~\cite{peng2021deep} and multiplex community detection~\cite{zhou2025community}. In contrast, TPA decomposes the model's \emph{internal} next-token distribution into explicit source contributions.

\paragraph{LLM-based Evaluation.}
External LLMs are also employed as verifiers. In RAG settings, outputs can be checked against retrieved evidence \cite{friel2023chainpoll} or through claim extraction and reference-based verification \cite{hu2024refchecker}, and LLM-as-a-judge baselines are often instantiated using curated prompts \cite{niu2024ragtruth}. Automated evaluation suites \cite{es2024ragas, trulens2024} have also been developed. Other strategies include cross-examination to expose inconsistencies \cite{cohen2023lm, yehuda2024interrogatellm} or fine-tuning detectors for span-level localization \cite{su2025learning}. Structured multi-agent frameworks have also been explored for domain-specific reasoning pipelines (e.g., legal consultation with statutory grounding)~\cite{lu2026querycounselstructuredreasoning}, where verifiers coordinate evidence retrieval and response refinement. However, many of these approaches require extra LLM calls or multi-step verification.

\begin{figure*}[!t]
    \centering
    \includegraphics[width=\linewidth]{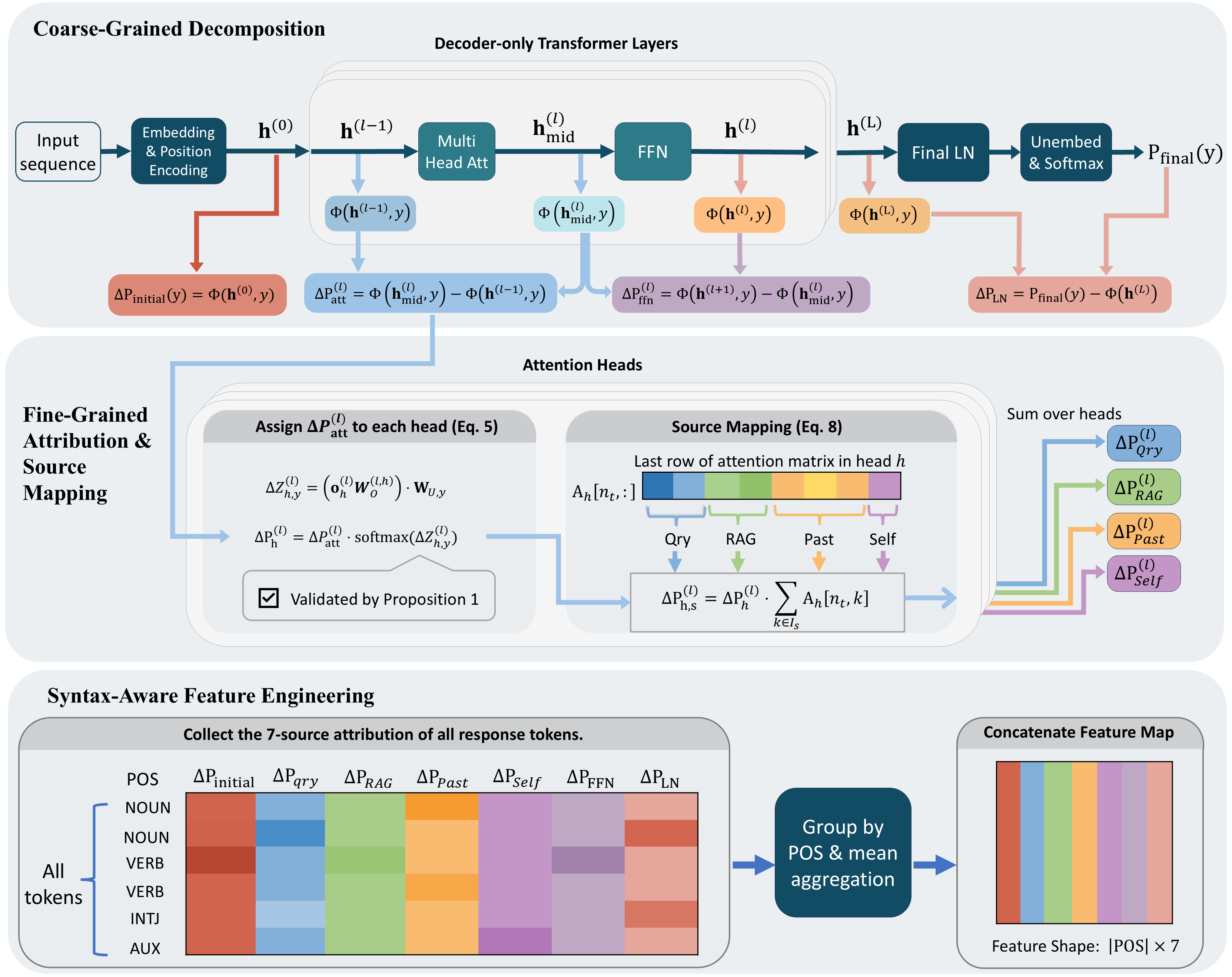}
\caption{\textbf{Overview of the TPA framework.} (1) \textbf{Coarse-Grained Decomposition}: Complete decomposition of token probability into four components (Section~\ref{sec:decomposition}). (2) \textbf{Fine-Grained Attribution}: Mapping attention contributions to four input sources via head-specific weights (Section~\ref{sec:attn_attr}). (3) \textbf{Syntax-Aware Feature Engineering}: Aggregating these attributions by POS tags to construct the final detection features (Section~\ref{sec:source-mapping}).}    \label{fig:framework}
\end{figure*}

\paragraph{Probing Internal Activations.} 
Recent work extracts factuality signals from internal representations, e.g., linear truthful directions or inference-time shifts \cite{burns2022discovering, li2023inference}, and probe-based detectors trained on hidden states \cite{azaria2023internal, han2024semantic}. Related studies show internal states remain predictive for hallucination detection \cite{chen2024inside}. Beyond detection, mechanistic analyses conflicts between FFN and RAG context \cite{sun2024redeep}, and lightweight indicators use attention-head norms \cite{novo}. Active approaches steer or edit activations \cite{tsv, li2023inference}, or adjust decoding probabilities for diagnosis \cite{chen2025attributive}. In contrast, we decompose the final token probability into fine-grained sources.

\section{Methodology}
\label{sec:methodology}

As illustrated in Figure~\ref{fig:framework}, TPA operates in three stages and {can be implemented} with a fully parallel teacher-forced pass. Given the generated response sequence $\mathbf{y}$ of length $T$, we can feed the entire sequence into the model with standard causal masking to extract hidden states and attention maps for all $T$ tokens in a single teacher-forced pass. This avoids autoregressive resampling while enabling efficient attribution computation.

We first derive a complete decomposition of token probabilities (Sec.~\ref{sec:decomposition}), then attribute attention contributions to specific attribution sources (Sec.~\ref{sec:attn_attr}). Finally, we aggregate these scores to quantify how sources drive distinct parts of speech (Sec.~\ref{sec:syntax_feature}). The pseudo-code and complexity analysis are provided in the Appendix. We report complexity instead of wall-clock time since the latter varies in different implementation hardware. To provide the theoretical basis for our method, we first formalize the transformer's architecture.

\subsection{Preliminaries: Transformer Architecture}
\subsubsection{Notations}
We consider a standard decoder-only Transformer with $L$ layers. We denote the query tokens as $\mathbf{x}_{\text{qry}}$, the retrieved context tokens as $\mathbf{x}_{\text{rag}}$, and the generated response as $\mathbf{y}=(y_1,\dots,y_T)$, with prompt length $T_0 = |\mathbf{x}_{\text{qry}}| + |\mathbf{x}_{\text{rag}}|$.
We analyze generation at step $t\in\{1,\dots,T\}$, where the model observes the prefix
$\mathbf{s}_t = [\mathbf{x}_{\text{qry}}, \mathbf{x}_{\text{rag}}, y_1,\dots,y_{t-1}],$
and predicts the next token $y_t$. Let
$n_t = |\mathbf{s}_t| = T_0 + t - 1 $
denote the {position index} of the last token in $\mathbf{s}_t$ (the token whose embedding is used to predict $y_t$).

Unless stated otherwise, all hidden states and residual outputs (e.g., $\mathbf{h}^{(l)}$, $\mathbf{h}^{(l)}_{\text{mid}}$) refer to the vector at the last position $n_t$, and we omit the explicit index $n_t$ and the step index $t$ for clarity. We keep explicit indices only for attention weights (e.g., $\mathbf{A}^{(l)}_h[n_t,k]$). We use $d$ for the hidden dimension, $H$ for the number of attention heads, and $d_h=d/H$ for the head dimension.

\subsubsection{Residual Updates and Probing}
The input tokens are mapped to continuous vectors via an embedding matrix $\mathbf{W}_e \in \mathbb{R}^{|\mathcal{V}| \times d}$ and summed with positional embeddings.
The initial state at the target position is $\mathbf{h}^{(0)} = \mathbf{W}_e[\mathbf{s}_t[n_t]] + \mathbf{p}_{n_t}$, where $\mathbf{p}_{n_t}$ is the positional embedding at position $n_t$.
We adopt the {Pre-LN} configuration. Crucially, each layer $l$ updates the hidden state via additive residual connections:
\begin{align}
\mathbf{h}_{\text{mid}}^{(l)} &= \mathbf{h}^{(l-1)} + \text{Attn}(\text{LN}(\mathbf{h}^{(l-1)})) \\
\mathbf{h}^{(l)} &= \mathbf{h}_{\text{mid}}^{(l)} + \text{FFN}(\text{LN}(\mathbf{h}_{\text{mid}}^{(l)}))
\end{align}
Here, $\text{Attn}(\cdot)$ denotes the attention output vector at position $n_t$ under causal masking.
This structure implies that the final representation is the sum of the initial embedding and all subsequent layer updates. To quantify these updates, we define a \textit{Probe Function} $\Phi(\mathbf{h}, y)$ similar to the logit lens technique \cite{logitlens} that measures the hypothetical probability of the target token $y$ given any intermediate state vector $\mathbf{h}$:
\begin{equation}\label{eq:prob}
\Phi(\mathbf{h}, y) = \left[\text{Softmax}(\mathbf{h} \mathbf{W}_{U})\right]_y
\end{equation}
where $\mathbf{W}_{U}$ is the unembedding matrix.

\noindent \textbf{Guiding Question:} \textit{Since the model is a stack of residual updates, can we mathematically decompose the final probability exactly into the sum of component contributions?}

\subsection{Coarse-Grained Decomposition}
\label{sec:decomposition}

We answer the preceding question affirmatively by leveraging the additive nature of the residual updates. Based on the probe function $\Phi(\mathbf{h}, y)$ defined in Eq. (\ref{eq:prob}), we isolate the probability contribution of each model component as the distinct change it induces in the probe output.

We define the baseline contribution from input static embeddings ($\Delta P_{\text{initial}}$), the incremental gains from Attention and FFN blocks in layer $l$ ($\Delta P_{\text{att}}^{(l)}, \Delta P_{\text{ffn}}^{(l)}$), and the adjustment from the final LayerNorm ($\Delta P_{\text{LN}}$) as follows:
\begin{align}
\Delta P_{\text{initial}}(y) &= \Phi(\mathbf{h}^{(0)}, y) \\
\Delta P_{\text{att}}^{(l)} &= \Phi(\mathbf{h}_{\text{mid}}^{(l)}, y) - \Phi(\mathbf{h}^{(l-1)}, y) \\
\Delta P_{\text{ffn}}^{(l)} &= \Phi(\mathbf{h}^{(l)}, y) - \Phi(\mathbf{h}_{\text{mid}}^{(l)}, y) \\
\Delta P_{\text{LN}} &= P_{\text{final}}(y) - \Phi(\mathbf{h}^{(L)}, y)
\end{align}
We define $P_{\text{final}}(y)$ as the model output probability after applying the final LayerNorm at position $n_t$.

By summing these differences, we derive the complete decomposition of the model's output.

\begin{theorem}[Complete Probability Decomposition]
\label{thm:decomposition_thm}
The final probability for a target token $y$ is exactly the sum of the contribution from the initial embedding, the cumulative contributions from Attention and FFN blocks across all $L$ layers, and the adjustment from the final LayerNorm:
\begin{equation}\label{eq:decomposition_thm}
\begin{aligned}
P_{\text{final}}(y) = & \Delta P_{\text{initial}}(y) + \Delta P_{\text{LN}} \\ & + \sum_{l=1}^{L} \left( \Delta P_{\text{att}}^{(l)} + \Delta P_{\text{ffn}}^{(l)} \right) \\ 
\end{aligned}
\end{equation}
\end{theorem}
\begin{proof}
    See Appendix.
\end{proof}

\noindent \textbf{The Guiding Question:} While Eq.~(\ref{eq:decomposition_thm}) quantifies \textit{how much} the model components contribute to the prediction probability, it treats the term $\Delta P_{\text{att}}^{(l)}$ as a black box. To effectively detect hallucinations, we must identify \textit{where} this attention is focused.

\subsection{Fine-Grained Attribution}
\label{sec:attn_attr}

To identify the focus of attention, we must decompose the attention contribution $\Delta P_{\text{att}}^{(l)}$ into contributions from individual attention heads.

\subsubsection{The Challenge: Non-Linearity}
Standard Multi-Head Attention concatenates the outputs of $H$ independent heads and projects them via an output matrix $\mathbf{W}_O^{(l)}$. Mathematically, by partitioning $\mathbf{W}_O^{(l)}$ into head-specific sub-matrices, this operation is strictly equivalent to the sum of projected head outputs:
\begin{equation}
\mathbf{h}_{\text{att}}^{(l)} = \sum_{h=1}^{H} \underbrace{\left(\mathbf{A}^{(l)}_h[n_t,:]\mathbf{V}^{(l)}_h\right)}_{\mathbf{o}^{(l)}_h} \mathbf{W}_O^{(l,h)}
\label{eq:mha_sum}
\end{equation}
where $\mathbf{o}^{(l)}_h$ is the head output vector at the target position, derived from the attention row $\mathbf{A}^{(l)}_h[n_t,:]$ and the value matrix $\mathbf{V}^{(l)}_h$.

Eq.~(\ref{eq:mha_sum}) establishes that the attention output is linear with respect to individual heads in the hidden state space. However, our goal is to attribute the {probability} change $\Delta P_{\text{att}}^{(l)}$ to each head $h$. Since the probe function $\Phi(\cdot)$ employs a non-linear Softmax operation, the sum of probability changes calculated by probing individual heads does not equal the attention block contribution.
This inequality prevents us from calculating head contributions by simply probing each head individually, motivating our shift to the logit space.

\subsubsection{Logit-Based Apportionment}
To bypass the non-linearity of the Softmax, we analyze contributions in the logit space. Let $\Delta z_{h,y}^{(l)}$ denote the scalar contribution of head $h$ to the logit of the \textbf{target token} $y$. This is calculated as the dot product between the projected head output and the target token's unembedding vector $\mathbf{w}_{U,y}$:
\begin{equation}
\Delta z_{h,y}^{(l)} = \left( \mathbf{o}^{(l)}_h \mathbf{W}_O^{(l,h)} \right) \cdot \mathbf{w}_{U,y}
\label{eq:head_logit_contrib}
\end{equation}
We then apportion the complete probability contribution $\Delta P_{\text{att}}^{(l)}$ (derived in Section \ref{sec:decomposition}) to each head $h$ proportional to its exponential logit contribution:
\begin{equation}
\Delta P_h^{(l)} = \Delta P_{\text{att}}^{(l)} \cdot \frac{\exp(\Delta z_{h,y}^{(l)})}{\sum_{j=1}^H \exp(\Delta z_{j,y}^{(l)})}
\label{eq:heuristic}
\end{equation}

\subsubsection{Theoretical Justification}
We ground the logit-based apportionment using a first-order Taylor expansion similar to \cite{montavon2019layer}. This approximates how logit changes affect the final probability.

\begin{proposition}[Linear Decomposition] \label{proposition}
The total attention contribution $\Delta P_{\text{att}}^{(l)}$ is approximated by the sum of head logits $\Delta z_{h,y}^{(l)}$ scaled by a gradient $\mathcal{G}^{(l)}$:
\begin{equation}
\Delta P_{\text{att}}^{(l)} \approx \mathcal{G}^{(l)} \sum_{h=1}^{H} \Delta z_{h,y}^{(l)} + \mathcal{E}
\end{equation}
\end{proposition}
\begin{proof} See Appendix. \end{proof}

While Proposition \ref{proposition} implies a linear relationship, direct attribution is unstable when head logits sum to zero. To resolve this, we employ the Softmax normalization in Eq.~(\ref{eq:heuristic}). This ensures numerical stability and constrains the sum of head scores to exactly match the layer total $\Delta P_{\text{att}}^{(l)}$. Thus, it preserves the conservation principle in Theorem \ref{thm:decomposition_thm}.

Regarding the approximation error $\mathcal{E}$, we rely on the first-order term for efficiency. This is effective because hallucinations are often high-confidence \citep{kadavath2022language}, which suppresses higher-order terms. For low-confidence scenarios, prior work identifies low probability itself as a strong hallucination signal \citep{guerreiro2023looking}. Our framework naturally captures this feature because the attribution scores sum exactly to the {final output probability} (Theorem \ref{thm:decomposition_thm}). Therefore, TPA inherently incorporates this critical probability signal and effectively detects such hallucinations.

\subsubsection{Source Mapping and The 7-Source Split}
\label{sec:source-mapping}
Having isolated the head contribution $\Delta P_h^{(l)}$, we can now answer the guiding question by tracing attention back to input tokens using the attention matrix $\mathbf{A}^{(l)}_h$.
We categorize inputs into four source types: $\mathcal{S} = \{\texttt{Qry}, \texttt{RAG}, \texttt{Past}, \texttt{Self}\}$.
Let $T_q = |\mathbf{x}_{\text{qry}}|$ and $T_r = |\mathbf{x}_{\text{rag}}|$, so $T_0 = T_q + T_r$ and $n_t = T_0 + t - 1$.
We partition the causal attention range $[1,n_t]$ into four disjoint index sets:
$\mathcal{I}_{\texttt{Qry}} = [1, T_q]$,
$\mathcal{I}_{\texttt{RAG}} = [T_q+1, T_0]$,
$\mathcal{I}_{\texttt{Past}} = [T_0+1, n_t-1]$,
and $\mathcal{I}_{\texttt{Self}} = \{n_t\}$.
For a source type $S$ containing token indices $\mathcal{I}_S$, the aggregated contribution is:
\begin{equation}
\Delta P_{S}^{(l)} = \sum_{h=1}^{H} \left( \Delta P_h^{(l)} \cdot \sum_{k \in \mathcal{I}_S} \mathbf{A}^{(l)}_{h}[n_t, k] \right)
\end{equation}

The seven sources are not ad hoc. The Pre-LN residual stream contributes three non-attention sources ($\Delta P_{\text{initial}}, \Delta P_{\text{ffn}}, \Delta P_{\text{LN}}$), and the attention contribution is partitioned into four sources ($\{\texttt{Qry}, \texttt{RAG}, \texttt{Past}, \texttt{Self}\}$) by the causal attention range into disjoint, exhaustive index sets. Coarser groupings (e.g., merging Past and Self) would collapse the current-vs-prior asymmetry that prior work links to RAG hallucination signals.

By aggregating these components, we achieve a complete partition of the final probability $P_{\text{final}}(y)$ into  {seven distinct sources}:
\begin{equation}
\begin{aligned}
P_{\text{final}}(y) &= \Delta P_{\text{initial}}(y) + \Delta P_{\text{LN}} \\
&\quad + \sum_{l=1}^{L} \left( \Delta P_{\text{ffn}}^{(l)} + \sum_{S \in \mathcal{S}} \Delta P_{S}^{(l)} \right)
\end{aligned}
\end{equation}

\noindent \textbf{The Guiding Question:} We have now derived a 7-dimensional attribution vector for every token. However, raw attribution scores lack context: a high FFN contribution might be normal for a function word but suspicious for a proper noun. How to contextualize these scores with syntactic priors?

\subsection{Syntax-Aware Feature Engineering}
\label{sec:syntax_feature}
To resolve this ambiguity, we employ Part-of-Speech (POS) tagging as a lightweight syntactic prior. Specifically, we assign a POS tag by Spacy \cite{spacy} to each generated token and aggregate the attribution scores for each grammatical category. By profiling which attribution sources (e.g., RAG) the LLM relies on for different parts of speech, we detect hallucination effectively.

\begin{table*}[!t]
\small
\centering
\setlength{\tabcolsep}{4.5pt}
\begin{tabular}{@{}l|ccc|ccc|ccc@{}}
\toprule
& \multicolumn{9}{c}{\textbf{RAGTruth}} \\ 
 \cmidrule(l){2-10}
\textbf{Method} & \multicolumn{3}{c|}{LLaMA2-7B} & \multicolumn{3}{c|}{LLaMA2-13B} & \multicolumn{3}{c}{LLaMA3-8B} \\ 
\cmidrule(r){2-4} \cmidrule(lr){5-7} \cmidrule(l){8-10}
\textbf{Metric} & \textbf{AUC} & \textbf{Recall} & \multicolumn{1}{c|}{\textbf{F1}} & \textbf{AUC} & \textbf{Recall} & \multicolumn{1}{c|}{\textbf{F1}} & \textbf{AUC} & \textbf{Recall} & \textbf{F1} \\ \midrule
ChainPoll~\cite{friel2023chainpoll} & 0.6738 & 0.7832 & \multicolumn{1}{c|}{0.7066} & 0.7414 & {0.7874} & \multicolumn{1}{c|}{0.7342} & 0.6687 & 0.4486 & 0.5813 \\
RAGAS~\cite{es2024ragas} & 0.7290 & 0.6327 & \multicolumn{1}{c|}{0.6667} & 0.7541 & 0.6763 & \multicolumn{1}{c|}{0.6747} & 0.6776 & 0.3909 & 0.5094 \\
Trulens~\cite{trulens2024} & 0.6510 & 0.6814 & \multicolumn{1}{c|}{0.6567} & 0.7073 & 0.7729 & \multicolumn{1}{c|}{0.6867} & 0.6464 & 0.3909 & 0.5053 \\
RefCheck~\cite{hu2024refchecker} & 0.6912 & 0.6280 & \multicolumn{1}{c|}{0.6736} & 0.7857 & 0.6800 & \multicolumn{1}{c|}{0.7023} & 0.6014 & 0.3580 & 0.4628 \\
EigenScore~\cite{chen2024inside} & 0.6045 & 0.7469 & \multicolumn{1}{c|}{0.6682} & 0.6640 & 0.6715 & \multicolumn{1}{c|}{0.6637} & 0.6497 & 0.7078 & 0.6745 \\
SEP~\cite{han2024semantic} & 0.7143 & 0.7477 & \multicolumn{1}{c|}{0.6627} & 0.8089 & 0.6580 & \multicolumn{1}{c|}{0.7159} & 0.7004 & 0.7333 & 0.6915 \\
ITI~\cite{li2023inference} & 0.7161 & 0.5416 & \multicolumn{1}{c|}{0.6745} & 0.8051 & 0.5519 & \multicolumn{1}{c|}{0.6838} & 0.6534 & 0.6850 & 0.6933 \\
ReDeEP~\cite{sun2024redeep} & 0.7458 & 0.8097 & \multicolumn{1}{c|}{\underline{0.7190}} & 0.8244 & 0.7198 & \multicolumn{1}{c|}{0.7587} & 0.7285 & \underline{0.7819} & 0.6947 \\
TSV~\cite{tsv} & 0.6609 & 0.5526 & \multicolumn{1}{c|}{0.6632} & 0.8123 & \textbf{0.8068} & \multicolumn{1}{c|}{0.6987} & 0.7769 & 0.5546 & 0.6442 \\
Novo~\cite{novo} & \underline{0.7608} & \underline{0.8274} & \multicolumn{1}{c|}{0.7057} & \underline{0.8506} & 0.7826 & \multicolumn{1}{c|}{\underline{0.7733}} & \textbf{0.8258} & 0.7737 & \underline{0.7801} \\
\midrule
\textbf{TPA} & \textbf{0.7873}$^\dagger$ & \textbf{0.8328} & \multicolumn{1}{c|}{\textbf{0.7238}$^\dagger$} & \textbf{0.8681}$^\dagger$ & \underline{0.7913} & \multicolumn{1}{c|}{\textbf{0.7975}$^\dagger$} & \underline{0.8211} & \textbf{0.7860} & \textbf{0.7843} \\ 
\toprule
& \multicolumn{9}{c}{\textbf{Dolly (AC)}} \\ 
 \cmidrule(l){2-10}
\textbf{Method} & \multicolumn{3}{c|}{LLaMA2-7B} & \multicolumn{3}{c|}{LLaMA2-13B} & \multicolumn{3}{c}{LLaMA3-8B} \\ 
\cmidrule(r){2-4} \cmidrule(lr){5-7} \cmidrule(l){8-10}
\textbf{Metric} & \textbf{AUC} & \textbf{Recall} & \multicolumn{1}{c|}{\textbf{F1}} & \textbf{AUC} & \textbf{Recall} & \multicolumn{1}{c|}{\textbf{F1}} & \textbf{AUC} & \textbf{Recall} & \textbf{F1} \\ \midrule
ChainPoll~\cite{friel2023chainpoll} & 0.3502 & 0.4138 & \multicolumn{1}{c|}{0.5581} & 0.4758 & 0.4364 & \multicolumn{1}{c|}{0.6000} & 0.2691 & 0.3415 & 0.4516 \\
RAGAS~\cite{es2024ragas} & 0.2877 & 0.5345 & \multicolumn{1}{c|}{0.6392} & 0.2840 & 0.4182 & \multicolumn{1}{c|}{0.5476} & {0.3628} & \underline{0.8000} & 0.5246 \\
Trulens~\cite{trulens2024} & 0.3198 & 0.5517 & \multicolumn{1}{c|}{0.6667} & 0.2565 & 0.3818 & \multicolumn{1}{c|}{0.4941} & 0.3352 & 0.3659 & 0.5172 \\
RefCheck~\cite{hu2024refchecker} & 0.2494 & 0.3966 & \multicolumn{1}{c|}{0.5412} & 0.2869 & 0.2545 & \multicolumn{1}{c|}{0.3944} & -0.0089 & 0.1951 & 0.2759 \\
EigenScore~\cite{chen2024inside} & 0.2428 & 0.7500 & \multicolumn{1}{c|}{0.7241} & 0.2948 & 0.8181 & \multicolumn{1}{c|}{0.7200} & 0.2065 & 0.7142 & 0.5952 \\
SEP~\cite{han2024semantic} & 0.2605 & 0.6216 & \multicolumn{1}{c|}{0.7023} & 0.2823 & 0.6545 & \multicolumn{1}{c|}{0.6923} & 0.0639 & 0.6829 & 0.6829 \\
ITI~\cite{li2023inference} & 0.0442 & 0.5816 & \multicolumn{1}{c|}{0.6281} & 0.0646 & 0.5385 & \multicolumn{1}{c|}{0.6712} & 0.0024 & 0.3091 & 0.4250 \\
ReDeEP~\cite{sun2024redeep} & 0.5136 & \underline{0.8245} & \multicolumn{1}{c|}{\textbf{0.7833}} & 0.5842 & \underline{0.8518} & \multicolumn{1}{c|}{\underline{0.7603}} & 0.3652 & \textbf{0.8392} & \underline{0.7100} \\
TSV~\cite{tsv} & \textbf{0.7454} & \textbf{0.8728} & \multicolumn{1}{c|}{\underline{0.7684}} & \underline{0.7552} & 0.5952 & \multicolumn{1}{c|}{0.6043} & {0.7347} & 0.6467 & 0.6695 \\
Novo~\cite{novo} & 0.6423 & 0.8070 & \multicolumn{1}{c|}{0.7244} & 0.6909 & 0.7222 & \multicolumn{1}{c|}{0.6903} & \underline{0.7418} & 0.5854 & 0.6316 \\
\midrule
\textbf{TPA} & \underline{0.7134} & 0.7897 & \multicolumn{1}{c|}{0.7527} & \textbf{0.8159}$^\dagger$ & \textbf{0.9741}$^\dagger$ & \multicolumn{1}{c|}{\textbf{0.8075}$^\dagger$} & \textbf{0.7608}$^\dagger$ & 0.6561 & \textbf{0.7529}$^\dagger$ \\ 
\bottomrule
\end{tabular}
\caption{Results on RAGTruth and Dolly (AC). TPA results are averaged over 5 random seeds. The dagger symbol $^\dagger$ indicates statistically significant improvement ($p < 0.05$) over the strongest baseline. Bold values indicate the best performance and underlined values indicate the second-best. Full results in Appendix.}
\label{table1}
\end{table*}

\subsubsection{Tag Propagation Strategy}
A mismatch problem arises because LLMs may split a single word into multiple tokens while standard POS taggers process whole words. We resolve this {granularity issue} via {tag propagation}: generated sub-word tokens inherit the POS tag of their parent word. For instance, if the noun "modification" is tokenized into [modi, fication], both sub-tokens are assigned the \texttt{NOUN} tag.

\subsubsection{Aggregation}
We first define the attribution vector $\mathbf{v}_t \in \mathbb{R}^7$ for each token $y_t$ as the concatenation of its 7 source components derived in Section \ref{sec:source-mapping}. Then we compute the mean attribution for each POS tag $\tau$:
\begin{equation}
\bar{\mathbf{v}}_\tau = \frac{\sum_{t: \text{POS}(y_t) = \tau} \mathbf{v}_t}{|\{t \mid \text{POS}(y_t) = \tau\}|} 
\end{equation}
The final feature vector $\mathbf{f} \in \mathbb{R}^{7 \times |\text{POS}|}$ is the concatenation of these POS-specific vectors. This feature combines source attribution with linguistic structure, forming a basis for hallucination detection.

\section{Experiments}

\subsection{Experimental Setup}\label{sec:setup}
We treat hallucination detection as a supervised binary classification task.\footnote{Code is available at \url{https://github.com/pengqian-lu/TPA}.} We employ an ensemble of 5 XGBoost \cite{xgboost} classifiers initialized with different random seeds. The input is a {126-dimensional syntax-aware feature vector}, constructed by aggregating the 7-source attribution scores across 18 universal POS tags (e.g., NOUN, VERB) defined by SpaCy \cite{spacy}.
For Qwen3-8B \citep{yang2025qwen3}, we generate new responses on RAGTruth's queries and retrieved contexts and use Claude Opus 4.6 as an LLM-as-judge to determine whether each response is hallucinated. We validate this judge on RAGTruth's 450 human-annotated Llama2-7B responses, where it attains 0.82 accuracy, 0.83 F1, and 0.64 Cohen's $\kappa$.
See more implementation details in Appendix~\ref{sec:impl_details}.

\begin{table}[t]
    \centering
    \small
    \begin{tabular}{lccc}
        \toprule
        \textbf{Method} & \textbf{F1} & \textbf{AUC} & \textbf{Recall} \\
        \midrule
        TSV \cite{tsv} & 0.6764 & 0.7972 & 0.5538 \\
        Novo \cite{novo} & 0.8000 & 0.8419 & 0.8765 \\
        \midrule
        \textbf{TPA} & \textbf{0.8702}$^\dagger$ & \textbf{0.9096}$^\dagger$ & \textbf{0.9200}$^\dagger$ \\
        \bottomrule
    \end{tabular}
    \caption{Performance comparison on the RAGTruth dataset using the Mistral-7B model. \textbf{Bold} indicates the best performance, and $^\dagger$ indicates statistically significant improvement over the strongest baseline ($p < 0.05$). The TPA results are averaged over 5 runs.}
    \label{tab:mistral_results}
\end{table}

\subsection{Dataset and Baselines}

To ensure a fair comparison, we utilize the public RAG hallucination benchmark established by \citet{sun2024redeep}. This benchmark consists of the Dolly (AC) and RAGTruth datasets. The former includes responses from Llama2 (7B/13B) and Llama3 (8B), while the latter covers these models in addition to Mistral-7B. Implementation details are provided in Appendix.
We compare our method against representative approaches from three categories introduced in Section~\ref{sec:related}. The introduction of baselines is provided in Appendix. For Mistral-7B, we compare against TSV and Novo. ReDeEP's Copying Heads were identified only on LLaMA-family backbones in the original release, with no head selection provided for Mistral. Other probing baselines such as SEP and ITI release training code but no Mistral-7B checkpoints, requiring full per-model probe or steering-direction re-fitting. Dolly does not include Mistral-7B responses, so Mistral is evaluated on RAGTruth only.

\subsection{Comparison with Baselines}\label{sec:baselines}

We evaluate TPA against baselines on RAGTruth (Llama2, Llama3, Mistral, Qwen3-8B) and Dolly (AC) (Llama2, Llama3). As shown in Table~\ref{table1} and Table~\ref{tab:mistral_results}, TPA is competitive across benchmarks.

On \textbf{RAGTruth}, TPA achieves statistically significant Rank-1 results ($p<0.05$) on Llama2-7B and Llama2-13B for both F1 and AUC. The largest improvement appears on Mistral-7B, where TPA reaches 0.8702 F1, outperforming Novo by 7\%, indicating good transfer to newer architectures with sliding-window attention. On Llama3-8B, TPA ranks first in F1 and Recall but is statistically comparable to the strongest baselines, suggesting a smaller margin on newer models. We also report results on Qwen3-8B \citep{yang2025qwen3} to verify generalization to a newer-generation backbone. As shown in Table~\ref{tab:qwen3_results}, TPA again outperforms Novo and TSV on F1, AUC, and Recall.

On \textbf{Dolly (AC)}, results show a scaling trend. TPA trails baselines (e.g., ReDeEP) on Llama2-7B, but becomes stronger as model capacity increases: it secures significant Rank-1 performance on Llama2-13B and the best F1 on Llama3-8B.

\begin{table}[t]
    \centering
    \small
    \begin{tabular}{lccc}
        \toprule
        \textbf{Method} & \textbf{F1} & \textbf{AUC} & \textbf{Recall} \\
        \midrule
        TSV~\cite{tsv}  & 0.5493 & 0.7710 & 0.6259 \\
        Novo~\cite{novo} & 0.5468 & 0.7919 & 0.6259 \\
        \midrule
        \textbf{TPA} & \textbf{0.6006}$^\dagger$ & \textbf{0.8236}$^\dagger$ & \textbf{0.7130}$^\dagger$ \\
        \bottomrule
    \end{tabular}
    \caption{Performance comparison on the RAGTruth dataset using the Qwen3-8B model. \textbf{Bold} indicates the best performance, and $^\dagger$ indicates statistically significant improvement over the strongest baseline ($p<0.05$). The TPA results are averaged over 5 runs.}
    \label{tab:qwen3_results}
\end{table}

\paragraph{Runtime.}\label{sec:runtime}
TPA's attribution pipeline adds three post-hoc stages (probe decomposition, head-wise attribution, source mapping) with total complexity $\mathcal{O}(L T |\mathcal{V}| d + L T d^2 + L H T^2)$. On an A100-40GB, feature extraction takes about 20 seconds per response in our sequential implementation ($\sim$17 GPU-hours per LLM per dataset). Classifier inference is under one second per response. See Appendix~\ref{sec:complexity} for the full derivation.

\begin{figure*}[t!]
    \centering
    \begin{subfigure}{0.48\linewidth}
        \centering
        \includegraphics[width=\linewidth]{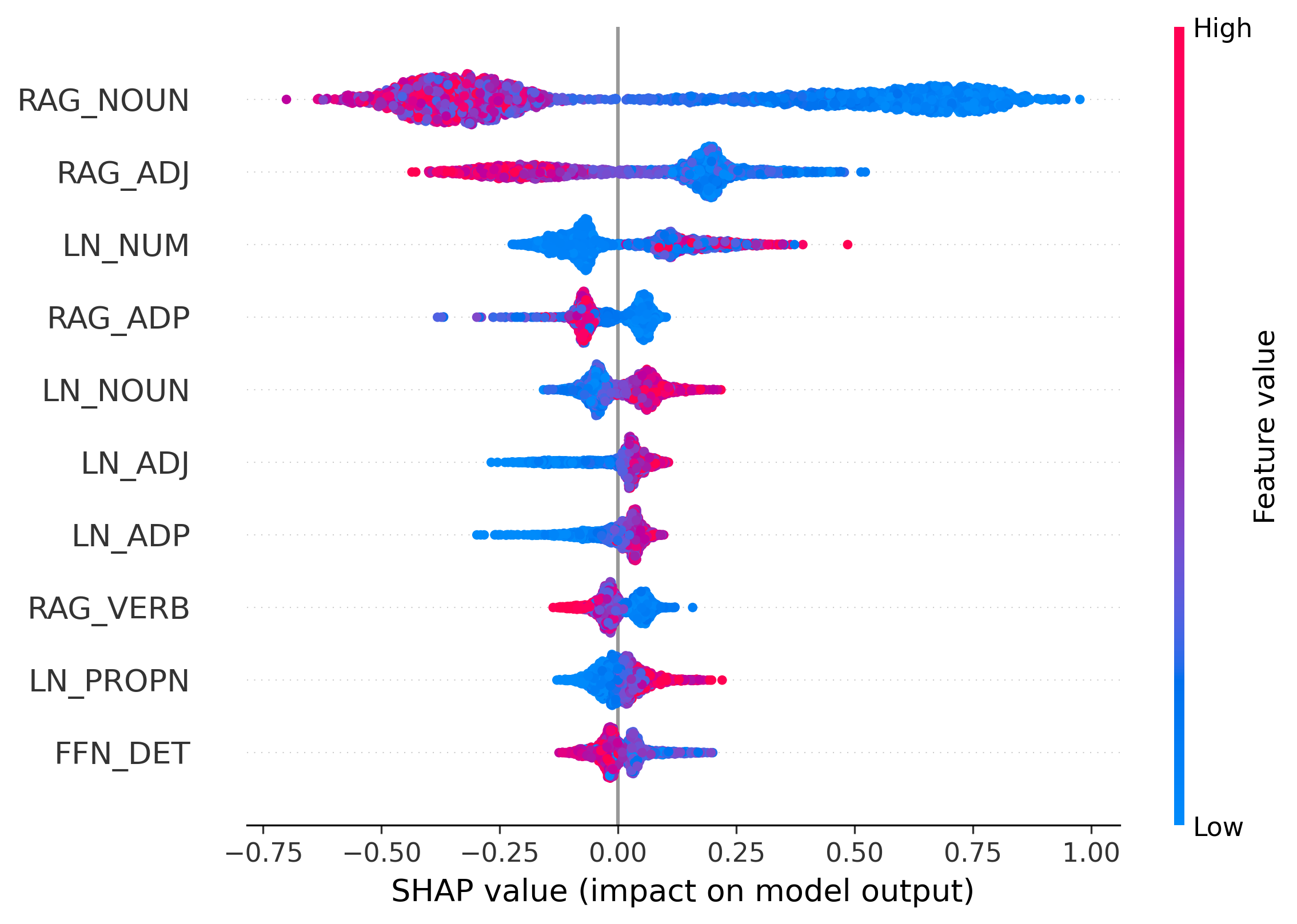}
        \caption{Llama2-7B}
        \label{fig:shap_7b}
    \end{subfigure}
    \hfill
    \begin{subfigure}{0.48\linewidth}
        \centering
        \includegraphics[width=\linewidth]{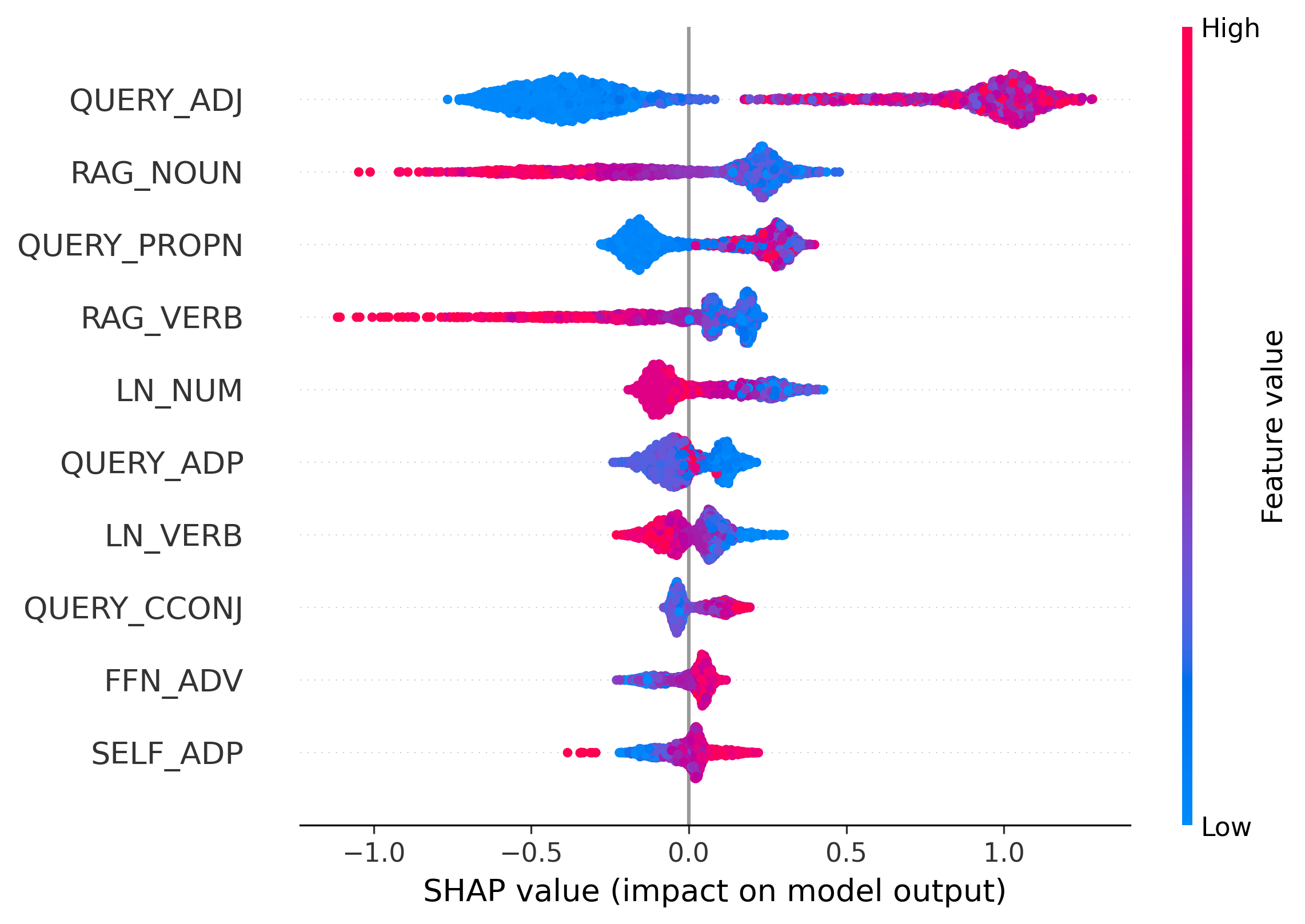}
        \caption{Llama2-13B}
        \label{fig:shap_13b}
    \end{subfigure}
    
    \caption{{SHAP summary plots illustrating the decision logic.} We visualize the top-10 features for classifiers trained on the {RAGTruth subsets corresponding to} Llama2-7B and Llama2-13B. Plots for Llama3-8B and Mistral-7B are provided in Appendix. The x-axis represents the SHAP value. Positive values indicate a push towards classifying the response as a {Hallucination}. The color represents the feature value (Red = High attribution, Blue = Low).}
    \label{fig:shap_analysis}   
\end{figure*}

\begin{figure}[t]
    \centering
    \includegraphics[width=\linewidth]{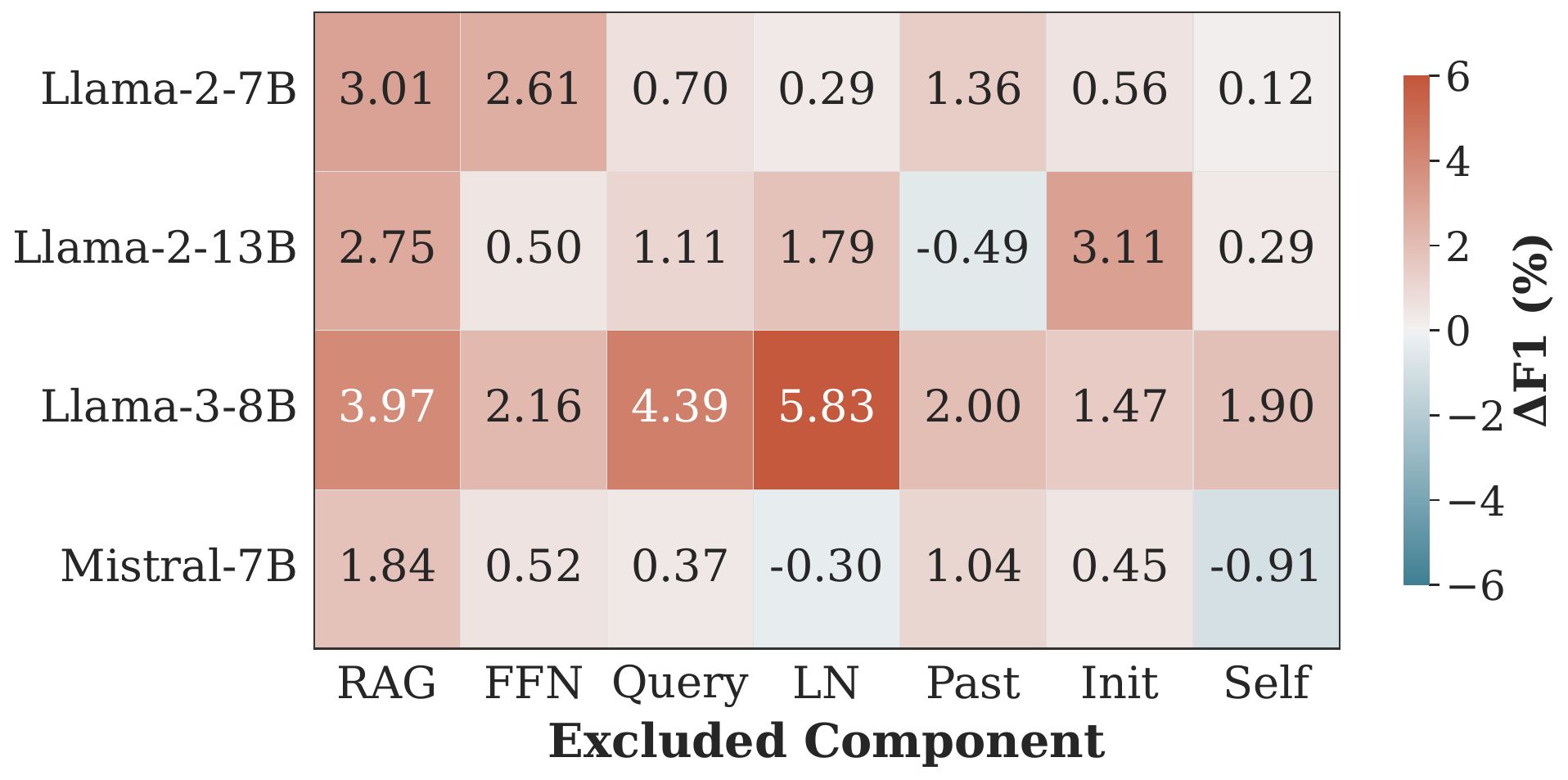}
    \caption{F1 Score Drop by Removing Components.}
    \label{fig:ablation}
\end{figure}

\section{Ablation Study Analysis}
We conduct an ablation study on RAGTruth to validate TPA (Figure~\ref{fig:ablation}). The full method generally achieves the best performance. Removing core components like \texttt{RAG} or \texttt{FFN} consistently degrades accuracy (e.g., a 3.01\% F1 drop on Llama-2-7B), confirming the importance of the retrieval-parameter conflict. Crucially, previously overlooked sources are distinctively vital. For instance, removing \texttt{LN} causes a sharp 5.83\% drop on Llama-3-8B. While excluding specific components yields marginal gains in several cases (e.g., \texttt{Self} on Mistral-7B), we retain the complete feature set to maintain a unified framework robust across diverse architectures. We exclude Dolly as its small sample size makes fine-grained feature evaluation unstable.

We further ablate the Part-of-Speech aggregation step in Section~\ref{sec:syntax_feature}. \textbf{TPA-POS} denotes our default features that aggregate the seven attribution sources separately within each POS tag. We compare it against two simpler aggregations that pool over all tokens regardless of syntactic category: \textbf{TPA-Mean} takes the mean of each source over all tokens in a response, yielding a 7-dimensional feature, while \textbf{TPA-Stat} additionally appends the per-source standard deviation, yielding 14 dimensions. As shown in Table~\ref{tab:pos_ablation} in the Appendix, TPA-POS attains the best F1 in all seven backbone-dataset settings, with gains of 1.8--4.6\% F1 over Mean on RAGTruth and up to 22.5\% on Dolly (LLaMA2-13B). Mean and Stat erase the source-POS interactions that the syntax-conditioned aggregation preserves.

\section{Interpretability Analysis} \label{sec:feature_analysis}
We apply SHAP analysis to classifiers trained on RAGTruth to validate our design principles. Results for Llama2 are shown in Figure~\ref{fig:shap_analysis}, while results for Llama3-8B and Mistral-7B are detailed in Appendix due to space constraints. We obtain three observations from this analysis.

\noindent \textbf{Observation 1: Fine-grained attribution is necessary.}
Relying solely on the binary conflict between internal FFN knowledge and external RAG context is insufficient for robust detection. As shown in Figure~\ref{fig:shap_analysis}, the classifier frequently depends on features derived from other components. For instance, the feature \texttt{LN\_NUM} plays a decisive role in Llama2-7B. This pattern indicates that the specific contribution from the Final LayerNorm to Numeral tokens serves as a critical signal. Similarly, query-based features like \texttt{QUERY\_NOUN} appear as top predictors in Mistral-7B. These findings confirm that accurate hallucination detection requires a complete decomposition of the generative process.

\noindent \textbf{Observation 2: Syntactic aggregation captures model-specific grounding logic.}
Different architectures ground information via distinct syntactic structures. While \texttt{RAG\_NOUN} dominates in Llama2 (Figure~\ref{fig:shap_analysis}) and Mistral, Llama3-8B relies heavily on relational structures, with \texttt{RAG\_ADP} (Adpositions) ranking highest. TPA's POS-based aggregation is thus essential to capture these diverse patterns.

\noindent \textbf{Observation 3: Hallucination fingerprints vary across architectures.}
Our analysis shows that hallucination signals are model-specific rather than universal. A clear example is \texttt{LN\_NUM}: high attribution is a strong hallucination signal in Llama2-7B, but this pattern reverses in Llama2-13B (Figure~\ref{fig:shap_13b}), where higher values correlate with factuality (negative SHAP values). This reversal suggests that larger models may use LayerNorm differently to regulate output distributions, motivating a learnable, syntax-aware detector.

\section{Conclusion and Future Work}
We introduced TPA to attribute token probability into seven distinct sources. By combining these with syntax-aware features, our framework effectively detects RAG hallucinations and outperforms baselines. Our results show that hallucination signals vary across models. This confirms the need for a learnable approach rather than static heuristics.


Future work will proceed in four directions. First, extending TPA to phrase- or span-level attribution for efficiency. Second, active mitigation via online monitoring and intervention on risky patterns (e.g., abnormal \texttt{FFN} or \texttt{LN} reliance). Third, studying TPA under noisy or adversarial retrieval, where expected source shifts may themselves serve as detection signals. Fourth, extending POS-based aggregation to non-English and specialized domains (code, scientific text) via multilingual spaCy or AST node types.

\section*{Limitations}
Our framework presents three limitations. First, it relies on {white-box access} to model internals, preventing application to closed-source APIs. Second, the decomposition incurs higher {computational overhead} than simple scalar probes. Third, our feature engineering depends on {external linguistic tools} (POS taggers), which may limit generalization to specialized domains like code generation where standard syntax is less defined.

\section*{Acknowledgements}
This work was supported by the Australian Research Council through the Laureate Fellow Project under Grant FL190100149.

\bibliography{bib}

\appendix
\section{Proof of Theorem 1}
We expand the right-hand side (RHS) of Eq.~\ref{eq:decomposition_thm} by substituting the definitions of each term. 

First, consider the summation term. By substituting $\Delta P_{\text{att}}^{(l)}$ and $\Delta P_{\text{ffn}}^{(l)}$, the intermediate probe value $\Phi(\mathbf{h}_{\text{mid}}^{(l)})$ cancels out within each layer:
\begin{equation}
\begin{aligned}
&\sum_{l=1}^{L} \left( \Delta P_{\text{att}}^{(l)} + \Delta P_{\text{ffn}}^{(l)} \right) \\
&= \sum_{l=1}^{L} ( [\Phi(\mathbf{h}_{\text{mid}}^{(l)}) \\& \quad - \Phi(\mathbf{h}^{(l-1)})] + [\Phi(\mathbf{h}^{(l)}) - \Phi(\mathbf{h}_{\text{mid}}^{(l)})] ) \\
&= \sum_{l=1}^{L} \left( \Phi(\mathbf{h}^{(l)}) - \Phi(\mathbf{h}^{(l-1)}) \right)
\end{aligned}
\notag
\end{equation}
This summation forms a telescoping series where adjacent terms cancel:
\begin{equation}
\begin{aligned}
& \sum_{l=1}^{L} (\Phi(\mathbf{h}^{(l)}) - \Phi(\mathbf{h}^{(l-1)})) \\ & = (\Phi(\mathbf{h}^{(1)}) - \Phi(\mathbf{h}^{(0)})) + \dots \\
& + (\Phi(\mathbf{h}^{(L)}) - \Phi(\mathbf{h}^{(L-1)})) \\
&= \Phi(\mathbf{h}^{(L)}) - \Phi(\mathbf{h}^{(0)})
\end{aligned}
\notag
\end{equation}
Now, substituting this result, along with the definitions of $\Delta P_{\text{initial}}$ and $\Delta P_{\text{LN}}$, back into the full RHS expression:
\begin{equation}
\begin{aligned}
& \text{RHS} = \underbrace{\Phi(\mathbf{h}^{(0)})}_{\Delta P_{\text{initial}}} + \underbrace{(P_{\text{final}} - \Phi(\mathbf{h}^{(L)}))}_{\Delta P_{\text{LN}}} \\ & + \underbrace{(\Phi(\mathbf{h}^{(L)}) - \Phi(\mathbf{h}^{(0)}))}_{\text{Summation}} = P_{\text{final}}
\end{aligned}
\notag
\end{equation}
The RHS simplifies exactly to $P_{\text{final}}(y)$, which completes the proof.

\section{Proof of Proposition 1}
\label{app:proof_prop1}

\begin{proof}
We consider the $l$-th layer of the Transformer. Let $\mathbf{h}^{(l-1)} \in \mathbb{R}^d$ be the input hidden state vector at the target position $n_t$.
The Multi-Head Attention mechanism computes a residual update $\Delta \mathbf{h}$ by summing the outputs of $H$ heads:
\begin{equation}
\Delta \mathbf{h} = \sum_{h=1}^{H} \mathbf{u}_h
\end{equation}
where $\mathbf{u}_h \in \mathbb{R}^d$ is the projected output of the $h$-th head.

The probe function $\Phi(\mathbf{h}, y)$ computes the probability of the target token $y$ by projecting the hidden state onto the vocabulary logits $\mathbf{z} \in \mathbb{R}^{|\mathcal{V}|}$ and applying the Softmax function:
\begin{equation}
\Phi(\mathbf{h}, y) = \frac{\exp(\mathbf{z}_y)}{\sum_{v\in\mathcal{V}} \exp(\mathbf{z}_v)}, \text{where } \mathbf{z}_v = \mathbf{h} \cdot \mathbf{w}_{U,v}
\end{equation}
Here, $\mathbf{w}_{U,v}$ is the unembedding vector for token $v$ from matrix $\mathbf{W}_{U}$. For brevity, let $p_y = \Phi(\mathbf{h}^{(l-1)}, y)$ denote the probability of the target token at the current state.

We approximate the change in probability, $\Delta P_{\text{att}}^{(l)}$, using a first-order Taylor expansion of $\Phi$ with respect to $\mathbf{h}$ around $\mathbf{h}^{(l-1)}$:
\begin{equation}
\begin{aligned}
\Delta P_{\text{att}}^{(l)} &= \Phi(\mathbf{h}^{(l-1)} + \Delta \mathbf{h}, y) - \Phi(\mathbf{h}^{(l-1)}, y) \\
&\approx \nabla_{\mathbf{h}} \Phi(\mathbf{h}^{(l-1)}, y)^\top \cdot \Delta \mathbf{h} \\
&= \sum_{h=1}^{H} \left( \nabla_{\mathbf{h}} \Phi(\mathbf{h}^{(l-1)}, y)^\top \cdot \mathbf{u}_h \right)
\end{aligned}
\label{eq:taylor_expansion}
\end{equation}

To compute the gradient $\nabla_{\mathbf{h}} \Phi$, we apply the chain rule through the logits $z_v$. The partial derivative of the Softmax output $p_y$ with respect to any logit $z_v$ is given by $p_y(\delta_{yv} - p_v)$, where $\delta$ is the Kronecker delta. The gradient of the logit $z_v$ with respect to $\mathbf{h}$ is simply $\mathbf{w}_{U,v}$. Thus:
\begin{equation}
\begin{aligned}
\nabla_{\mathbf{h}} \Phi &= \sum_{v\in\mathcal{V}} \frac{\partial \Phi}{\partial z_v} \frac{\partial z_v}{\partial \mathbf{h}} \\
&= \sum_{v\in\mathcal{V}} p_y(\delta_{yv} - p_v) \mathbf{w}_{U,v} \\
&= p_y(1 - p_y)\mathbf{w}_{U,y} - \sum_{v \neq y} p_y p_v \mathbf{w}_{U,v}
\end{aligned}
\end{equation}

Substituting this gradient back into Eq.~(\ref{eq:taylor_expansion}) for a specific head contribution term (denoted as $\text{Term}_h$):
\begin{equation}
\begin{aligned}
\text{Term}_h &= \nabla_{\mathbf{h}} \Phi^\top \cdot \mathbf{u}_h \\
&= \underbrace{p_y(1 - p_y)}_{\mathcal{G}^{(l)}} (\mathbf{w}_{U,y}^\top \cdot \mathbf{u}_h) - \\&\quad \underbrace{\sum_{v \neq y} p_y p_v (\mathbf{w}_{U,v}^\top \cdot \mathbf{u}_h)}_{\mathcal{E}_h}
\end{aligned}
\end{equation}
We observe that the dot product $\mathbf{w}_{U,y}^\top \cdot \mathbf{u}_h$ is strictly equivalent to the scalar logit contribution $\Delta z_{h,y}^{(l)}$ defined in Eq. \ref{eq:head_logit_contrib}.
The factor $\mathcal{G}^{(l)} = p_y(1 - p_y)$ represents the gradient common to all heads, depending only on the layer input $\mathbf{h}^{(l-1)}$.
Therefore, the contribution of head $h$ is dominated by the linear term $\mathcal{G}^{(l)} \cdot \Delta z_{h,y}^{(l)}$, subject to the off-target error term $\mathcal{E}_h$.
\end{proof}
\begin{algorithm*}[!t]
\caption{TPA Part I: Token Probability Attribution (Teacher-Forced Pass / sequential process)}
\label{alg:TPA_attribution}
\begin{algorithmic}[1]
\REQUIRE Transformer $\mathcal{M}$, Query Tokens $\mathbf{x}_{\text{qry}}$, Retrieved Context Tokens $\mathbf{x}_{\text{rag}}$
\ENSURE Sequence of 7-source attribution vectors $\mathcal{V} = (\mathbf{v}_1, \dots, \mathbf{v}_T)$

\STATE Initialize contexts $\{\mathbf{C}_t\}_{t=1}^{T}$ and attribution storage $\mathcal{V} \leftarrow \emptyset$
\STATE $\mathbf{C}_1 \leftarrow [\mathbf{x}_{\text{qry}}, \mathbf{x}_{\text{rag}}]$
\FOR{position $t = 2 \dots T$}
    \STATE $\mathbf{C}_t \leftarrow [\mathbf{C}_{t-1}, y_{t-1}]$
\ENDFOR

\FOR{position $t = 1 \dots T$}
    \STATE \textbf{1. Forward Pass \& State Caching}
    \STATE Run the model on $\mathbf{C}_t$ and cache states at the last position $n_t = |\mathbf{C}_t|$.
    \STATE Cache $\mathbf{h}^{(0)}$, $\mathbf{h}_{\text{mid}}^{(l)}$, $\mathbf{h}^{(l)}$, and Attention Maps $\mathbf{A}^{(l)}$ for all layers $l \in [1, L]$.
    
    \STATE \textbf{2. Coarse Decomposition (Residual Stream)}
    \STATE Initialize $\mathbf{v}_t \in \mathbb{R}^7$ with zeros.
    \STATE $\mathbf{v}_t[\texttt{Init}] \leftarrow \text{Probe}(\mathbf{h}^{(0)}, y_t)$ \COMMENT{Contribution from initial embedding}
    \STATE $\mathbf{v}_t[\texttt{LN}] \leftarrow P_{\text{final}}(y_t) - \text{Probe}(\mathbf{h}^{(L)}, y_t)$ \COMMENT{Contribution from final LayerNorm}
    
    \STATE \textbf{3. Layer-wise Attribution}
    \FOR{layer $l = 1$ to $L$}
        \STATE $\mathbf{v}_t[\texttt{FFN}] \mathrel{+}= \text{Probe}(\mathbf{h}^{(l)}, y_t) - \text{Probe}(\mathbf{h}_{\text{mid}}^{(l)}, y_t)$
        
        \STATE $\Delta P_{\text{att}} \leftarrow \text{Probe}(\mathbf{h}_{\text{mid}}^{(l)}, y_t) - \text{Probe}(\mathbf{h}^{(l-1)}, y_t)$
        
        \STATE \textbf{4. Fine-Grained Decomposition (Head \& Source)}
        \FOR{head $h = 1$ to $H$}
            \STATE Let $\mathbf{o}_h$ be the output vector of head $h$.
            \STATE Compute logit update: $\Delta z_h \leftarrow \left(\mathbf{o}_h \mathbf{W}_O^{(l,h)}\right)\cdot \mathbf{w}_{U,y_t}$
            \STATE Compute ratio $\omega_h \leftarrow \frac{\exp(\Delta z_{h})}{\sum_{j} \exp(\Delta z_{j})}$ \COMMENT{Logit-based apportionment}
            \STATE $\Delta P_h \leftarrow \Delta P_{\text{att}} \times \omega_h$
            
            \FOR{source $S \in \{\texttt{Qry}, \texttt{RAG}, \texttt{Past}, \texttt{Self}\}$}
            \STATE Sum attention weights on indices of $S$: $a_{h,S} \leftarrow \sum_{k \in \mathcal{I}_S} \mathbf{A}^{(l)}_h[n_t, k]$
            \STATE $\mathbf{v}_t[S] \mathrel{+}= \Delta P_h \cdot a_{h,S}$
            \ENDFOR
        \ENDFOR
    \ENDFOR
    
    \STATE Store $\mathbf{v}_t$ in attribution matrix $\mathcal{V}$.
\ENDFOR
\RETURN $\mathcal{V}, \text{Generated Tokens } \mathbf{y}$
\end{algorithmic}
\end{algorithm*}

\begin{algorithm*}[!t]
\caption{TPA Part II: Syntax-Aware Feature Aggregation with Sub-word Tag Propagation}
\label{alg:TPA_aggregation}
\begin{algorithmic}[1]
\REQUIRE Generated Tokens $\mathbf{y} = (y_1, \dots, y_T)$, Attribution Vectors $\mathcal{V} = (\mathbf{v}_1, \dots, \mathbf{v}_T)$
\ENSURE Syntax-Aware Feature Vector $\mathbf{f} \in \mathbb{R}^{126}$

\STATE \textbf{1. String Reconstruction \& Alignment Map}
\STATE Decode tokens $\mathbf{y}$ into a complete string \texttt{text}.
\STATE Construct an alignment map $M$ where $M[i]$ contains the list of token indices corresponding to the $i$-th word in \texttt{text}.

\STATE \textbf{2. POS Tagging \& Propagation}
\STATE Initialize tag list $\mathcal{T}$ of length $T$.
\STATE Run POS tagger (e.g., SpaCy) on string \texttt{text} to obtain words $W_1, \dots, W_K$ and tags $\text{tag}_1, \dots, \text{tag}_K$.

\FOR{each word index $k = 1$ to $K$}
    \STATE Get corresponding token indices: $\mathcal{I}_{\text{tokens}} \leftarrow M[k]$
    \STATE Get POS tag for the word: $c \leftarrow \text{tag}_k$
    \FOR{each token index $t \in \mathcal{I}_{\text{tokens}}$}
        \STATE $\tau_t \leftarrow c$ \COMMENT{Propagate parent word's tag to all sub-word tokens}
    \ENDFOR
\ENDFOR

\STATE \textbf{3. Syntax-Aware Aggregation}
\STATE Initialize feature vector $\mathbf{f} \leftarrow \emptyset$.
\STATE Define set of Universal POS tags $\mathcal{P}_{\text{univ}}$.

\FOR{each POS category $c \in \mathcal{P}_{\text{univ}}$}
    \STATE Identify tokens belonging to this category: $\mathcal{I}_c = \{t \mid \tau_t = c\}$
    
    \IF{$\mathcal{I}_c \neq \emptyset$}
        \STATE \textit{// Compute mean attribution profile for this syntactic category}
        \STATE $\bar{\mathbf{v}}_c \leftarrow \frac{1}{|\mathcal{I}_c|} \sum_{t \in \mathcal{I}_c} \mathbf{v}_t$
    \ELSE
        \STATE $\bar{\mathbf{v}}_c \leftarrow \mathbf{0}_7$ \COMMENT{Fill with zeros if category is absent in response}
    \ENDIF
    
    \STATE $\mathbf{f} \leftarrow \text{Concatenate}(\mathbf{f}, \bar{\mathbf{v}}_c)$
\ENDFOR

\RETURN $\mathbf{f}$
\end{algorithmic}
\end{algorithm*}

\section{Implementation Details}
\label{sec:impl_details}

\paragraph{Environment and Models.}
All experiments were conducted on a computational node equipped with a single NVIDIA A100 (40GB) GPU and 200GB of RAM.
Our software stack uses CUDA 12.8, Python 3.10, PyTorch 2.x, and HuggingFace Transformers 4.56.1.
We evaluate our framework using three Large Language Models: Llama2-7b-chat, Llama2-13b-chat~\cite{llama2}, and Llama3-8b-instruct~\cite{llama3}.

Due to GPU memory constraints (40GB), we implement TPA using a sequential prefix-replay procedure (token-by-token), rather than a fully parallel teacher-forced pass.
On our hardware, generating the full TPA feature vector for one response takes approximately 20 seconds on average.
Since each dataset contains fewer than 3{,}000 samples, the total feature-extraction cost is on the order of $\sim$17 GPU-hours per dataset (per evaluated LLM).

Hallucination detection is performed using an ensemble of five XGBoost classifiers; inference is typically well below one second per response, and the total classification cost per dataset is only a few CPU-hours, negligible compared to feature extraction.

For POS tagging, we use spaCy with \texttt{en\_core\_web\_sm} and disable NER.

\paragraph{Feature Extraction and Classifier.}
For each response, we extract the 7-dimensional attribution vector for every token and aggregate them based on 18 universal POS tags (e.g., NOUN, VERB) defined by the SpaCy library \cite{spacy}. This results in a fixed-size feature vector ($7 \times 18 = 126$ dimensions) for each sample.

\paragraph{Training and Evaluation Protocols.}
To ensure fair comparison and statistical robustness, we tailor our training strategies to the data availability of each dataset. We implement strict data isolation to prevent leakage. Crucially, to mitigate the variance inherent in small-data scenarios, we adopt a Multi-Seed Ensemble Strategy. For every experiment, we repeat the entire evaluation process using 5 distinct outer random seeds. For each seed, we construct an ensemble of 5 XGBoost classifiers. The final prediction is derived via Hard Voting (majority rule) for binary classification metrics (F1, Recall) and Soft Voting (probability averaging) for AUC.

\noindent \textbf{Protocol I: Standard Split (RAGTruth Llama2-7b/13b/Mistral).}
For datasets with official splits, we utilize the standard training and test sets.
\begin{itemize}
    \item \textbf{Optimization:} We employ \textbf{Optuna} with a TPE sampler to optimize hyperparameters. We run 50 trials maximizing the F1-score using 5-fold Stratified Cross-Validation on the training set.
    \item \textbf{Training:} For each of the 5 outer seeds, we train an ensemble of 5 models. Each ensemble member is trained on a distinct 85\%/15\% split of the training data to facilitate diversity and Early Stopping (patience=50).
    \item \textbf{Evaluation:} Predictions are aggregated via voting on the held-out test set.
\end{itemize}

\noindent \textbf{Protocol II: Stratified 20-Fold CV (RAGTruth Llama3-8b).}
As the Llama3-8b subset lacks a training split, we adopt a Stratified 20-Fold Cross-Validation.
\begin{itemize}
    \item \textbf{Optimization:} Hyperparameters are optimized via Optuna on the available data prior to the cross-validation loop.
    \item \textbf{Training:} We iterate through 20 folds. Within each fold, we train the 5-member XGBoost ensemble on the training partition (using diverse internal splits for early stopping).
    \item \textbf{Aggregation:} Predictions from all folds are concatenated to compute the final performance metrics for each outer seed.
\end{itemize}

\noindent \textbf{Protocol III: Nested Leave-One-Out CV (Dolly).}
Given the limited size of the Dolly dataset ($N=100$), we implement a rigorous Nested Leave-One-Out (LOO) Cross-Validation.
\begin{itemize}
    \item \textbf{Outer Loop:} We iterate 100 times, isolating a single sample for testing in each iteration.
    \item \textbf{Inner Loop (Optimization):} On the remaining 99 samples, we conduct independent hyperparameter searches using Optuna (50 trials).
    \item \textbf{Ensemble Training:} For each LOO step, we train 5 XGBoost models on the 99 training samples. To handle class imbalance, we dynamically adjust the \texttt{scale\_pos\_weight} parameter.
    \item \textbf{Inference:} The final verdict for the single test sample is determined by the hard vote of the 5 ensemble members. This process is repeated for all 5 outer seeds to verify statistical significance.
\end{itemize}

\paragraph{Implementation Note regarding Memory Constraints.}
 While TPA is theoretically designed for single-pass parallel execution via teacher forcing, storing the full attention matrices $\mathcal{O}(T^2)$ and computational graphs for long sequences can be memory-intensive. In our specific experiments, due to GPU memory limitations (NVIDIA A100 40G), we implemented the attribution process sequentially (token-by-token). We emphasize that this implementation is mathematically equivalent to the parallel version due to the causal masking mechanism of Transformer decoders. The choice between serial and parallel implementation represents a trade-off between efficiency and memory usage, without affecting the attribution values or detection performance reported in this paper.

\paragraph{Hyperparameter Search and Best Value Discussion.}
We utilize the Optuna framework with a Tree-structured Parzen Estimator (TPE) sampler to perform automated hyperparameter tuning. For each model and data split, we run 50 trials to maximize the F1-score. The comprehensive search space is presented in Table~\ref{tab:hyperparams}.
Regarding the best-found values, our analysis reveals that the optimal configuration is highly dependent on the specific LLM and dataset size. We observed a consistent preference for moderate tree depths ($4 \le \text{max\_depth} \le 6$) and stronger regularization ($\lambda \ge 1.5$, $\gamma \ge 0.2$) across most experiments, indicating that preventing overfitting is critical given the high dimensionality of our feature space relative to the dataset size. Conversely, the optimal learning rate varied significantly ($0.01$ to $0.1$) depending on the base model (e.g., Llama-2 vs. Llama-3). Therefore, rather than fixing a single set of hyperparameters, we adopt a dynamic optimization strategy where the best parameters are re-evaluated for each fold and seed. This approach ensures that our reported results reflect the robust performance of the method rather than a specific tuning artifact.

\begin{table}[h]
\centering
\small
\caption{Hyperparameter search space for the XGBoost classifier in TPA.}
\label{tab:hyperparams}
\begin{tabular}{@{}ll@{}}
\toprule
\textbf{Hyperparameter} & \textbf{Search Values} \\ \midrule
Learning Rate & \{0.01, 0.02, 0.05, 0.1\} \\
Max Depth & \{4, 5, 6, 7\} \\
Subsample & \{0.6, 0.7, 0.8\} \\
Colsample By Tree & \{0.7, 0.8, 0.9\} \\
Gamma ($\gamma$) & \{0.1, 0.2, 0.5\} \\
Reg Alpha ($\alpha$) & \{0.01, 0.1, 0.5\} \\
Reg Lambda ($\lambda$) & \{1, 1.5, 2\} \\ \midrule
\textit{Fixed Parameters} & n\_estimators=1000, patience=50 \\
\bottomrule
\end{tabular}
\end{table}

\paragraph{Qwen3-8B Setup.}
Responses are generated with greedy decoding (\texttt{temperature=0.0}, \texttt{enable\_thinking=False}) using Qwen3's native chat template without a system prompt, matching RAGTruth's decoding configuration. TPA attribution is computed with a separate fork of HuggingFace Transformers 4.55.0 (the version with native Qwen3 support) that applies the Pre-LN probe and per-head source decomposition to \texttt{Qwen3Attention} and \texttt{Qwen3DecoderLayer}. We use the same 126-dimensional feature vector (18 POS $\times$ 7 contributions) and 5-seed $\times$ 5-ensemble XGBoost protocol as for Llama. Because the Qwen3-8B response set on RAGTruth is class-imbalanced toward the negative (no-hallucination) label, we set \texttt{scale\_pos\_weight} to the negative/positive ratio on the training set to maintain comparable class balance during optimization.

\begin{table}[h]
    \centering
    \small
    \setlength{\tabcolsep}{4pt}
    \begin{tabular}{lccc}
        \toprule
        \textbf{Backbone} & \textbf{TPA-POS} & \textbf{TPA-Mean} & \textbf{TPA-Stat} \\
        \midrule
        \multicolumn{4}{l}{\textit{RAGTruth}} \\
        LLaMA2-7B     & \textbf{0.7238} & 0.7055 & 0.7162 \\
        LLaMA2-13B    & \textbf{0.7975} & 0.7513 & 0.7645 \\
        LLaMA3-8B     & \textbf{0.7843} & 0.7458 & 0.7541 \\
        Mistral-7B    & \textbf{0.8702} & 0.8490 & 0.8521 \\
        \midrule
        \multicolumn{4}{l}{\textit{Dolly (AC)}} \\
        LLaMA2-7B     & \textbf{0.7527} & 0.6769 & 0.7206 \\
        LLaMA2-13B    & \textbf{0.8075} & 0.5828 & 0.6009 \\
        LLaMA3-8B     & \textbf{0.7529} & 0.5947 & 0.5869 \\
        \bottomrule
    \end{tabular}
    \caption{F1 score of TPA with different aggregation strategies (5-seed average). TPA-POS wins in all seven settings.}
    \label{tab:pos_ablation}
\end{table}

\paragraph{Artifacts and intended use.}
We use publicly available benchmarks (RAGTruth and Dolly (AC)) and open-access LLM checkpoints strictly for research evaluation, consistent with their intended research use.
We do not redistribute any original datasets or model weights; our released artifact is research code and documentation for reproducing the experiments, and it instructs users to obtain the datasets/models from their official sources.

\section{Baselines Introduction}
\begin{enumerate}
    \item \textbf{EigenScore/INSIDE}\cite{chen2024inside} Focuses on detecting hallucination by evaluating the response's semantic consistency, defined as the log-determinant of the covariance matrix of the LLM's internal states during response generation.

    \item \textbf{SEP}\cite{han2024semantic} Proposes a linear model to detect hallucination based on semantic entropy at test time without requiring multiple responses.

    \item \textbf{SAPLMA}\cite{azaria2023internal} Detecting hallucination based on the hidden layer activations of LLMs.

    \item \textbf{ITI}\cite{li2023inference} Detecting hallucination based on the hidden layer activations of LLMs.

    \item \textbf{Ragtruth Prompt}\cite{niu2024ragtruth} Provides prompts for an LLM-as-judge to detect hallucination in the RAG setting.


    \item \textbf{ChainPoll}\cite{friel2023chainpoll} Provides prompts for an LLM-as-judge to detect hallucination in the RAG setting.

    \item \textbf{RAGAS}\cite{es2024ragas} Uses an LLM to split the response into a set of statements and verify whether each statement is supported by the retrieved documents. If any statement is not supported, the response is considered hallucinated.

    \item \textbf{Trulens}\cite{trulens2024} Evaluating the overlap between the retrieved documents and the generated response to detect hallucination by a LLM.

    \item \textbf{P(True)}\cite{kadavath2022language} The paper detects hallucinations by having the model estimate the probability that its own generated answer is correct, based on the key assumption that it is often easier for a model to recognize a correct answer than to generate one.

    \item \textbf{SelfCheckGPT}\cite{manakul2023selfcheckgpt} SelfCheckGPT detects hallucinations by checking for informational consistency across multiple stochastically sampled responses, based on the assumption that factual knowledge leads to consistent statements while hallucinations lead to divergent and contradictory ones.

    \item \textbf{LN-Entropy}\cite{malinin2021structured} This paper detects hallucinations by quantifying knowledge uncertainty, which it measures primarily with a novel metric called Reverse Mutual Information that captures the disagreement across an ensemble's predictions, with high RMI indicating a likely hallucination.

    \item \textbf{Energy}\cite{liu2020energy} This paper detects hallucinations by using an energy score, derived directly from the model's logits, as a more reliable uncertainty measure than softmax confidence to identify out-of-distribution inputs that cause the model to hallucinate.

    \item \textbf{Focus}\cite{zhang2023enhancing} This paper detects hallucinations by calculating an uncertainty score focused on keywords, and then refines it by propagating penalties from unreliable context via attention and correcting token probabilities using entity types and inverse document frequency to mitigate both overconfidence and underconfidence.

    \item \textbf{Perplexity}\cite{renout} This paper detects hallucinations by separately measuring the Relative Mahalanobis Distance for both input and output embeddings, based on the assumption that in-domain examples will have embeddings closer to their respective foreground (in-domain) distributions than to a generic background distribution.
    
    \item \textbf{REFCHECKER}\cite{hu2024refchecker} Uses an LLM to extract claim triplets from a response and verifies them with another LLM to detect hallucination.

    \item \textbf{REDEEP}\cite{sun2024redeep} Detects hallucination by analyzing the balance between contributions from Copying Heads that process external context and Knowledge FFNs that inject internal knowledge, based on the finding that RAG hallucinations often arise from conflicts between these two sources. This method has two versions: token level and chunk level. We compare with the latter since it generally has better performance.

    \item \textbf{NoVo}\cite{novo} It leverages the L2 norms of specific attention heads as reliable indicators of truthfulness. By identifying a subset of truth-correlated heads from a small reference set, it employs a voting mechanism based on these head norms to detect hallucinations without requiring model parameter updates.
    
    \item \textbf{TSV}\cite{tsv} It introduces a lightweight steering vector to reshape the LLM's latent space during inference. By actively intervening to enhance the linear separability between truthful and hallucinated representations in the hidden states, it enables effective detection using a simple classifier on the steered embeddings.
\end{enumerate}

\section{Complexity Analysis of the Attribution Process}
\label{sec:complexity}

In this section, we rigorously analyze the computational overhead of our attribution framework. 
We report complexity in terms of analysis passes and asymptotic costs, as wall-clock varies substantially with caching strategies and kernel implementations.
We focus strictly on the attribution extraction process for a generated response of length $T$. Let $L$, $d$, $|\mathcal{V}|$, and $H$ denote the number of layers, hidden dimension, vocabulary size, and attention heads (per layer), respectively.
The standard inference complexity for a Transformer is $\mathcal{O}(L \cdot T \cdot d^2 + L \cdot H \cdot T^2)$. Our attribution process introduces post-hoc computations, decomposed into three specific stages:

\paragraph{1. Complete Probability Decomposition.}
To satisfy Theorem 1, we must compute the complete probability changes using the probe function $\Phi(\mathbf{h}, y)$. The bottleneck is the calculation of the global partition function (denominator) in Softmax.
\begin{itemize}
    \item \textbf{Mechanism:} The probe function $\Phi(\mathbf{h}, y) = \text{Softmax}(\mathbf{h}\mathbf{W}_U)_y = \frac{\exp(\mathbf{w}_{U,y}^\top \mathbf{h})}{\sum_{v\in\mathcal{V}} \exp(\mathbf{w}_{U,v}^\top \mathbf{h})}$ requires projecting the hidden state $\mathbf{h}$ to the full vocabulary logits $\mathbf{z} = \mathbf{h}\mathbf{W}_U$.
    \item \textbf{Single Probe Complexity:} For a single hidden state $\mathbf{h} \in \mathbb{R}^d$, the matrix-vector multiplication with the unembedding matrix $\mathbf{W}_U \in \mathbb{R}^{d \times |\mathcal{V}|}$ costs $\mathcal{O}(|\mathcal{V}| \cdot d)$.
    \item \textbf{Total Calculation:} We must apply this probe at multiple points:
    \begin{enumerate}
        \item \textbf{Global Components:} For $\Delta P_{\text{initial}}$ and $\Delta P_{\text{LN}}$, the probe is called once per generation step. Cost: $\mathcal{O}(T \cdot |\mathcal{V}| \cdot d)$.
        \item \textbf{Layer Components:} For $\Delta P_{\text{att}}^{(l)}$ and $\Delta P_{\text{ffn}}^{(l)}$, the probe is invoked twice per layer (before and after the residual update). Summing over $L$ layers, this costs $\mathcal{O}(L \cdot T \cdot |\mathcal{V}| \cdot d)$.
    \end{enumerate}
    \item \textbf{Stage Complexity:} Combining these terms, the dominant complexity is $\mathcal{O}(L \cdot T \cdot |\mathcal{V}| \cdot d)$.
\end{itemize}

\paragraph{2. Head-wise Attribution.}
Once $\Delta P_{\text{att}}^{(l)}$ is obtained, we apportion it to individual heads based on their contribution to the target logit.
\begin{itemize}
    \item \textbf{Mechanism:} This attribution requires projecting the target token vector $\mathbf{w}_{U,y}$ back into the hidden state space using the layer's output projection matrix $\mathbf{W}_O^{(l)} \in \mathbb{R}^{d \times d}$.
    \item \textbf{Step Complexity:} The calculation proceeds in two sub-steps:
    \begin{enumerate}
        \item \textbf{Projection:} We compute the projected target vector $\mathbf{g} = (\mathbf{W}_O^{(l)})^\top \mathbf{w}_{U,y}$. Since $\mathbf{W}_O^{(l)}$ is a $d \times d$ matrix, this matrix-vector multiplication costs $\mathcal{O}(d^2)$.
        \item \textbf{Assignment:} We distribute the contribution to $H$ heads by performing dot products between the head outputs $\mathbf{o}_h$ and the corresponding segments of $\mathbf{g}$. For $H$ heads, this sums to $\mathcal{O}(d)$.
    \end{enumerate}
    \item \textbf{Stage Complexity:} The projection step ($\mathcal{O}(d^2)$) dominates the assignment step ($\mathcal{O}(d)$). Integrating over $L$ layers and $T$ tokens, the total complexity is $\mathcal{O}(L \cdot T \cdot d^2)$.
\end{itemize}

\paragraph{3. Mapping Attention to Input Sources.}
Finally, we map head contributions to the four sources by aggregating attention weights $\mathbf{A} \in \mathbb{R}^{H \times |\mathbf{s}| \times |\mathbf{s}|}$. This involves two distinct sub-steps for each generated token at step $t$ within a single layer:
\begin{itemize}
    \item \textbf{Step 1: Summation.} For each head $h$, we sum the attention weights corresponding to specific source indices (e.g., $\mathcal{I}_{\texttt{RAG}}$):
    \begin{equation*}
        w_{h, S} = \sum_{k \in \mathcal{I}_S} \mathbf{A}_{h}[n_t, k]
    \end{equation*}
    This requires iterating over the causal range up to $n_t$. For $H$ heads, the cost is $\mathcal{O}(H \cdot n_t)$.
    \item \textbf{Step 2: Normalization \& Weighting.} We calculate the final source contribution by weighting the head contributions:
    \begin{equation*}
        \Delta P_{S} = \sum_{h=1}^H \Delta P_h \cdot \frac{w_{h,S}}{\sum_{\text{all sources}} w_{h,\cdot}}
    \end{equation*}
    This involves scalar operations proportional to the number of heads $H$. Cost: $\mathcal{O}(H)$.
    \item \textbf{Stage Complexity:} The summation step ($\mathcal{O}(H \cdot n_t)$) dominates. We sum this cost across all $L$ layers, and then accumulate over the generation steps $t=1$ to $T$. The calculation is $\sum_{t=1}^{T} (L \cdot H \cdot n_t) \approx \mathcal{O}(L \cdot H \cdot T^2)$.
\end{itemize}

\paragraph{Overall Efficiency.}
The total computational cost is the sum of these three stages:
\begin{equation*}
    \mathcal{C}_{\text{total}} = \mathcal{O}(\underbrace{L \cdot T \cdot |\mathcal{V}| \cdot d}_{\text{Prob. Decomp.}} + \underbrace{L \cdot T \cdot d^2}_{\text{Head Attr.}} + \underbrace{L \cdot H \cdot T^2}_{\text{Source Map}})
\end{equation*}

\paragraph{Runtime Efficiency.}
It is worth noting that theoretical complexity does not directly equate to wall-clock latency. Standard text generation is \textit{serial} (token-by-token), which limits GPU parallelization. 
In contrast, our framework {can} process the full sequence of length $T$ in a single parallel teacher-forced pass, enabling efficient GPU matrix operations. {When implemented this way}, it avoids the $K$ sequential generation passes required by baselines like SelfCheckGPT.

\begin{table*}[!t]
\small
\centering
\setlength{\tabcolsep}{3pt}
\begin{tabular}{@{}l|ccc|ccc|ccc@{}}
\toprule
& \multicolumn{9}{c}{\textbf{RAGTruth}} \\ 
 \cmidrule(l){2-10}
\textbf{Model} & \multicolumn{3}{c|}{LLaMA2-7B} & \multicolumn{3}{c|}{LLaMA2-13B} & \multicolumn{3}{c}{LLaMA3-8B} \\ 
\cmidrule(r){2-4} \cmidrule(lr){5-7} \cmidrule(l){8-10}
\textbf{Metric} & \textbf{AUC} & \textbf{Recall} & \multicolumn{1}{c|}{\textbf{F1}} & \textbf{AUC} & \textbf{Recall} & \multicolumn{1}{c|}{\textbf{F1}} & \textbf{AUC} & \textbf{Recall} & \textbf{F1} \\ \midrule
SelfCheckGPT \cite{manakul2023selfcheckgpt} & — & 0.4642 & \multicolumn{1}{c|}{0.4642} & — & 0.4642 & \multicolumn{1}{c|}{0.4642} & — & 0.5111 & 0.5111 \\
Perplexity \cite{renout} & 0.5091 & 0.5190 & \multicolumn{1}{c|}{0.6749} & 0.5091 & 0.5190 & \multicolumn{1}{c|}{0.6749} & 0.6235 & 0.6537 & 0.6778 \\
LN-Entropy \cite{malinin2021structured} & 0.5912 & 0.5383 & \multicolumn{1}{c|}{0.6655} & 0.5912 & 0.5383 & \multicolumn{1}{c|}{0.6655} & 0.7021 & 0.5596 & 0.6282 \\
Energy \cite{liu2020energy} & 0.5619 & 0.5057 & \multicolumn{1}{c|}{0.6657} & 0.5619 & 0.5057 & \multicolumn{1}{c|}{0.6657} & 0.5959 & 0.5514 & 0.6720 \\
Focus \cite{zhang2023enhancing} & 0.6233 & 0.5309 & \multicolumn{1}{c|}{0.6622} & 0.7888 & 0.6173 & \multicolumn{1}{c|}{0.6977} & 0.6378 & 0.6688 & 0.6879 \\
Prompt \cite{niu2024ragtruth} & — & 0.7200 & \multicolumn{1}{c|}{0.6720} & — & 0.7000 & \multicolumn{1}{c|}{0.6899} & — & 0.4403 & 0.5691 \\
ChainPoll \cite{friel2023chainpoll} & 0.6738 & 0.7832 & \multicolumn{1}{c|}{0.7066} & 0.7414 & {0.7874} & \multicolumn{1}{c|}{0.7342} & 0.6687 & 0.4486 & 0.5813 \\
RAGAS \cite{es2024ragas} & 0.7290 & 0.6327 & \multicolumn{1}{c|}{0.6667} & 0.7541 & 0.6763 & \multicolumn{1}{c|}{0.6747} & 0.6776 & 0.3909 & 0.5094 \\
Trulens \cite{trulens2024} & 0.6510 & 0.6814 & \multicolumn{1}{c|}{0.6567} & 0.7073 & 0.7729 & \multicolumn{1}{c|}{0.6867} & 0.6464 & 0.3909 & 0.5053 \\
RefCheck \cite{hu2024refchecker} & 0.6912 & 0.6280 & \multicolumn{1}{c|}{0.6736} & 0.7857 & 0.6800 & \multicolumn{1}{c|}{0.7023} & 0.6014 & 0.3580 & 0.4628 \\
P(True) \cite{kadavath2022language} & 0.7093 & 0.5194 & \multicolumn{1}{c|}{0.5313} & 0.7998 & 0.5980 & \multicolumn{1}{c|}{0.7032} & 0.6323 & 0.7083 & 0.6835 \\
EigenScore \cite{chen2024inside} & 0.6045 & 0.7469 & \multicolumn{1}{c|}{0.6682} & 0.6640 & 0.6715 & \multicolumn{1}{c|}{0.6637} & 0.6497 & 0.7078 & 0.6745 \\
SEP \cite{han2024semantic} & 0.7143 & 0.7477 & \multicolumn{1}{c|}{0.6627} & 0.8089 & 0.6580 & \multicolumn{1}{c|}{0.7159} & 0.7004 & 0.7333 & 0.6915 \\
SAPLMA \cite{azaria2023internal} & 0.7037 & 0.5091 & \multicolumn{1}{c|}{0.6726} & 0.8029 & 0.5053 & \multicolumn{1}{c|}{0.6529} & 0.7092 & 0.5432 & 0.6718 \\
ITI \cite{li2023inference} & 0.7161 & 0.5416 & \multicolumn{1}{c|}{0.6745} & 0.8051 & 0.5519 & \multicolumn{1}{c|}{0.6838} & 0.6534 & 0.6850 & 0.6933 \\
ReDeEP \cite{sun2024redeep} & 0.7458 & 0.8097 & \multicolumn{1}{c|}{\underline{0.7190}} & 0.8244 & 0.7198 & \multicolumn{1}{c|}{0.7587} & 0.7285 & \underline{0.7819} & 0.6947 \\
TSV \cite{tsv} & 0.6609 & 0.5526 & \multicolumn{1}{c|}{0.6632} & 0.8123 & \textbf{0.8068} & \multicolumn{1}{c|}{0.6987} & 0.7769 & 0.5546 & 0.6442 \\
Novo \cite{novo} & \underline{0.7608} & \underline{0.8274} & \multicolumn{1}{c|}{0.7057} & \underline{0.8506} & 0.7826 & \multicolumn{1}{c|}{\underline{0.7733}} & \textbf{0.8258} & 0.7737 & \underline{0.7801} \\
\midrule
\textbf{TPA} & \textbf{0.7873}$^\dagger$ & \textbf{0.8328} & \multicolumn{1}{c|}{\textbf{0.7238}$^\dagger$} & \textbf{0.8681}$^\dagger$ & \underline{0.7913} & \multicolumn{1}{c|}{\textbf{0.7975}$^\dagger$} & \underline{0.8211} & \textbf{0.7860} & \textbf{0.7843} \\ 
\textbf{Std} & \scriptsize{0.0007} & \scriptsize{0.0145} & \multicolumn{1}{c|}{\scriptsize{0.0039}} & \scriptsize{0.0075} & \scriptsize{0.0086} & \multicolumn{1}{c|}{\scriptsize{0.0076}} & \scriptsize{0.0025} & \scriptsize{0.0068} & \scriptsize{0.0053} \\
\textbf{P-val} & \scriptsize{${<}0.001$} & \scriptsize{${=}0.227$} & \multicolumn{1}{c|}{\scriptsize{${=}0.025$}} & \scriptsize{${=}0.003$} & \scriptsize{-} & \multicolumn{1}{c|}{\scriptsize{${=}0.001$}} & \scriptsize{-} & \scriptsize{${=}0.125$} & \scriptsize{${=}0.074$} \\
\toprule
& \multicolumn{9}{c}{\textbf{Dolly (AC)}} \\ 
 \cmidrule(l){2-10}
\textbf{Model} & \multicolumn{3}{c|}{LLaMA2-7B} & \multicolumn{3}{c|}{LLaMA2-13B} & \multicolumn{3}{c}{LLaMA3-8B} \\ 
\cmidrule(r){2-4} \cmidrule(lr){5-7} \cmidrule(l){8-10}
\textbf{Metric} & \textbf{AUC} & \textbf{Recall} & \multicolumn{1}{c|}{\textbf{F1}} & \textbf{AUC} & \textbf{Recall} & \multicolumn{1}{c|}{\textbf{F1}} & \textbf{AUC} & \textbf{Recall} & \textbf{F1} \\ \midrule
SelfCheckGPT \cite{manakul2023selfcheckgpt} & — & 0.1897 & \multicolumn{1}{c|}{0.3188} & 0.2728 & 0.1897 & \multicolumn{1}{c|}{0.3188} & 0.1095 & 0.2195 & 0.3600 \\
Perplexity \cite{renout} & 0.2728 & 0.7719 & \multicolumn{1}{c|}{0.7097} & 0.2728 & 0.7719 & \multicolumn{1}{c|}{0.7097} & 0.1095 & 0.3902 & 0.4571 \\
LN-Entropy \cite{malinin2021structured} & 0.2904 & 0.7368 & \multicolumn{1}{c|}{0.6772} & 0.2904 & 0.7368 & \multicolumn{1}{c|}{0.6772} & 0.1150 & 0.5365 & 0.5301 \\
Energy \cite{liu2020energy} & 0.2179 & 0.6316 & \multicolumn{1}{c|}{0.6261} & 0.2179 & 0.6316 & \multicolumn{1}{c|}{0.6261} & -0.0678 & 0.4047 & 0.4440 \\
Focus \cite{zhang2023enhancing} & 0.3174 & 0.5593 & \multicolumn{1}{c|}{0.6534} & 0.1643 & 0.7333 & \multicolumn{1}{c|}{0.6168} & 0.1266 & 0.6918 & 0.6874 \\
Prompt \cite{niu2024ragtruth} & — & 0.3965 & \multicolumn{1}{c|}{0.5476} & — & 0.4182 & \multicolumn{1}{c|}{0.5823} & — & 0.3902 & 0.5000 \\
ChainPoll \cite{friel2023chainpoll} & 0.3502 & 0.4138 & \multicolumn{1}{c|}{0.5581} & 0.4758 & 0.4364 & \multicolumn{1}{c|}{0.6000} & 0.2691 & 0.3415 & 0.4516 \\
RAGAS \cite{es2024ragas} & 0.2877 & 0.5345 & \multicolumn{1}{c|}{0.6392} & 0.2840 & 0.4182 & \multicolumn{1}{c|}{0.5476} & {0.3628} & \underline{0.8000} & 0.5246 \\
Trulens \cite{trulens2024} & 0.3198 & 0.5517 & \multicolumn{1}{c|}{0.6667} & 0.2565 & 0.3818 & \multicolumn{1}{c|}{0.4941} & 0.3352 & 0.3659 & 0.5172 \\
RefCheck \cite{hu2024refchecker} & 0.2494 & 0.3966 & \multicolumn{1}{c|}{0.5412} & 0.2869 & 0.2545 & \multicolumn{1}{c|}{0.3944} & -0.0089 & 0.1951 & 0.2759 \\
P(True) \cite{kadavath2022language} & 0.1987 & 0.6350 & \multicolumn{1}{c|}{0.6509} & 0.2009 & 0.6180 & \multicolumn{1}{c|}{0.5739} & 0.3472 & 0.5707 & 0.6573 \\
EigenScore \cite{chen2024inside} & 0.2428 & 0.7500 & \multicolumn{1}{c|}{0.7241} & 0.2948 & 0.8181 & \multicolumn{1}{c|}{0.7200} & 0.2065 & 0.7142 & 0.5952 \\
SEP \cite{han2024semantic} & 0.2605 & 0.6216 & \multicolumn{1}{c|}{0.7023} & 0.2823 & 0.6545 & \multicolumn{1}{c|}{0.6923} & 0.0639 & 0.6829 & 0.6829 \\
SAPLMA \cite{azaria2023internal} & 0.0179 & 0.5714 & \multicolumn{1}{c|}{0.7179} & 0.2006 & 0.6000 & \multicolumn{1}{c|}{0.6923} & -0.0327 & 0.4040 & 0.5714 \\
ITI \cite{li2023inference} & 0.0442 & 0.5816 & \multicolumn{1}{c|}{0.6281} & 0.0646 & 0.5385 & \multicolumn{1}{c|}{0.6712} & 0.0024 & 0.3091 & 0.4250 \\
ReDeEP \cite{sun2024redeep} & 0.5136 & \underline{0.8245} & \multicolumn{1}{c|}{\textbf{0.7833}} & 0.5842 & \underline{0.8518} & \multicolumn{1}{c|}{\underline{0.7603}} & 0.3652 & \textbf{0.8392} & \underline{0.7100} \\
TSV \cite{tsv} & \textbf{0.7454} & \textbf{0.8728} & \multicolumn{1}{c|}{\underline{0.7684}} & \underline{0.7552} & 0.5952 & \multicolumn{1}{c|}{0.6043} & {0.7347} & 0.6467 & 0.6695 \\
Novo \cite{novo} & 0.6423 & 0.8070 & \multicolumn{1}{c|}{0.7244} & 0.6909 & {0.7222} & \multicolumn{1}{c|}{0.6903} & \underline{0.7418} & 0.5854 & 0.6316 \\
\midrule
\textbf{TPA} & \underline{0.7134} & 0.7897 & \multicolumn{1}{c|}{0.7527} & \textbf{0.8159}$^\dagger$ & \textbf{0.9741}$^\dagger$ & \multicolumn{1}{c|}{\textbf{0.8075}$^\dagger$} & \textbf{0.7608} & 0.6561 & \textbf{0.7529}$^\dagger$ \\ 
\textbf{Std} & \scriptsize{0.0215} & \scriptsize{0.0227} & \multicolumn{1}{c|}{\scriptsize{0.0199}} & \scriptsize{0.0210} & \scriptsize{0.0096} & \multicolumn{1}{c|}{\scriptsize{0.0137}} & \scriptsize{0.0164} & \scriptsize{0.0452} & \scriptsize{0.0337} \\ 
\textbf{P-val} & \scriptsize{-} & \scriptsize{-} & \multicolumn{1}{c|}{\scriptsize{-}} & \scriptsize{${<}0.001$} & \scriptsize{${<}0.001$} & \multicolumn{1}{c|}{\scriptsize{${<}0.001$}} & \scriptsize{0.0025} & \scriptsize{-} & \multicolumn{1}{c|}{\scriptsize{${=}0.001$}} \\ \bottomrule
\end{tabular}
\caption{Full Results on RAGTruth and Dolly (AC) datasets. TPA results are reported as the Mean and Standard Deviation over 5 random seeds, obtained using an ensemble of 5 XGBoost models. $^\dagger$ indicates that the improvement over the strongest baseline is statistically significant ($p < 0.05$) under a one-sample t-test. P-values are reported for metrics where TPA achieves Rank 1; otherwise, a dash (-) is shown. Bold values indicate the best performance and underlined values indicate the second-best.}\label{fulltable1}
\end{table*}

\begin{figure*}[t!]
    \centering
    \begin{subfigure}{0.48\linewidth}
        \centering
        \includegraphics[width=\linewidth]{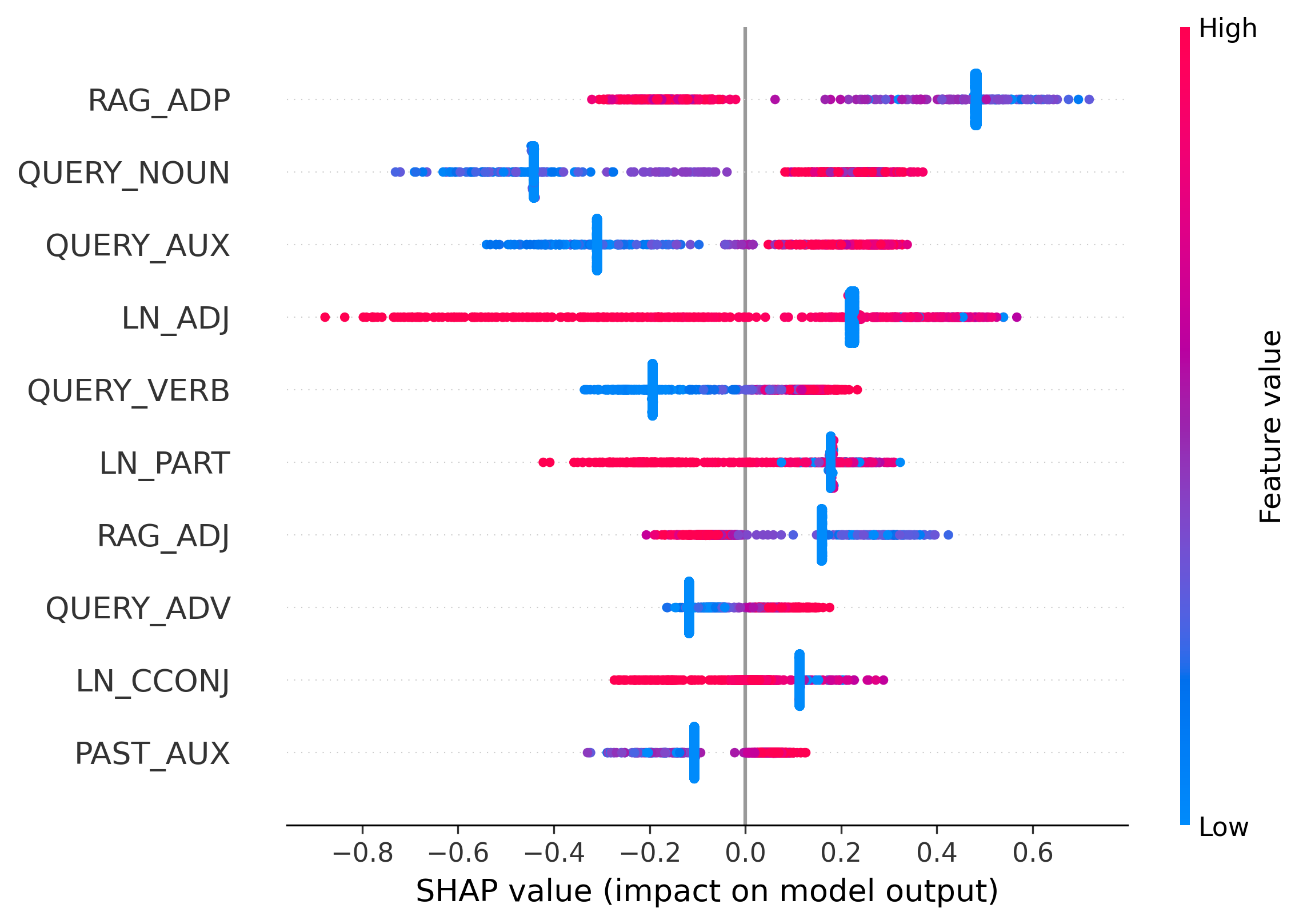}
        \caption{Llama3-8B}
        \label{fig:shap_8b}
    \end{subfigure}
    \hfill
    \begin{subfigure}{0.48\linewidth}
        \centering
        \includegraphics[width=\linewidth]{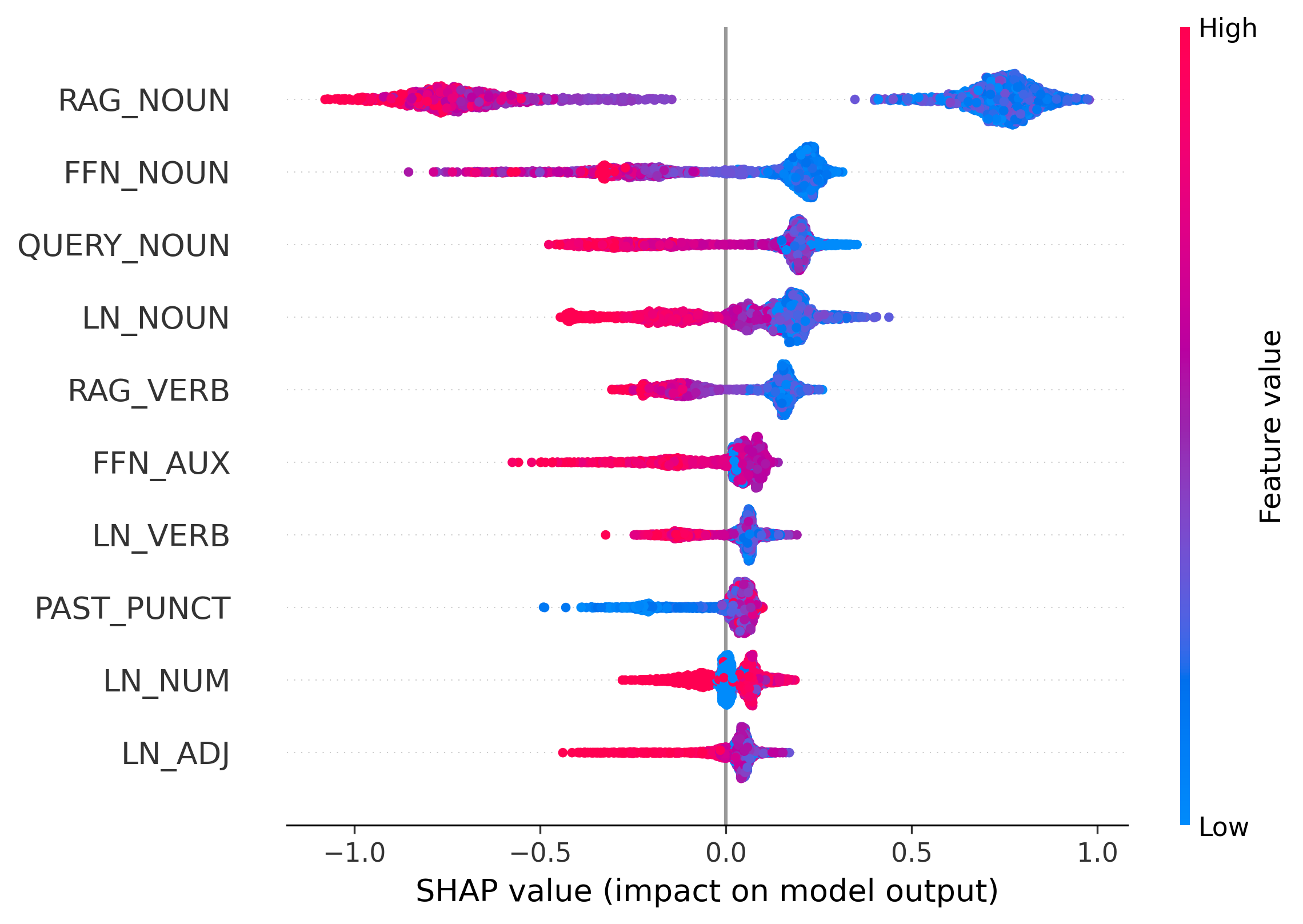}
        \caption{Mistral-7B}
        \label{fig:shap_mistral}
    \end{subfigure}
    
    \caption{{SHAP summary plots illustrating the decision logic.} We visualize the top-10 features for classifiers trained on the {RAGTruth subsets corresponding to}  Llama3-8B and Mistral-7B. The x-axis represents the SHAP value. Positive values indicate a push towards classifying the response as a {Hallucination}. The color represents the feature value (Red = High attribution, Blue = Low).}
    \label{fig:shap_analysis_appendix}
\end{figure*}

\section{AI Assistance Disclosure}
We used AI-based tools to assist with language editing and draft refinement. All technical content, experiments, and conclusions were produced and verified by the authors.

\end{document}